\documentclass[11pt,a4paper,logo]{googledeepmind}

\setleftlogo[180pt]{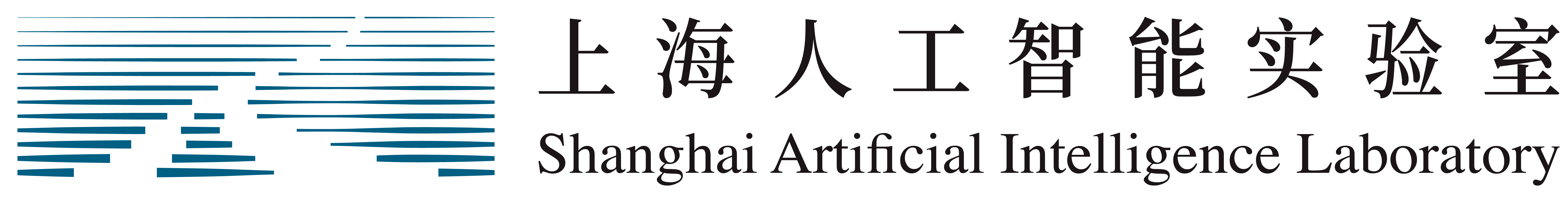} 
\setrightlogo[180pt]{}

\usepackage{placeins}
\usepackage{subcaption}

\usepackage[
    natbib=true,
    backend=biber,
    style=numeric, 
    sorting=none 
]{biblatex}
\AtEveryBibitem{\clearfield{month}}
\AtEveryBibitem{\clearfield{day}}

\usepackage{csquotes}

\newcommand{\ProjectName}{Agents-A1}
\newcommand{\github}{\raisebox{-1.5pt}{\includegraphics[height=1em]{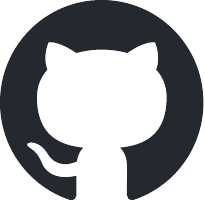}}}

\title{Scaling the Horizon, Not the Parameters: Reaching Trillion-Parameter Performance with a 35B Agent}

\author[1]{\ProjectName~Team, Shanghai Artificial Intelligence Laboratory \\
}

\usepackage{pdflscape}
\usepackage{amsthm}
\newtheorem{definition}{Definition}
\usepackage{textcomp}
\usepackage{rotating}
\usepackage{setspace}
\usepackage{soul} 
\usepackage{multicol}
\usepackage{subcaption}
\usepackage{caption}

\sethlcolor{green!14}

\usepackage{array}
\usepackage{multirow}
\usepackage[justification=centering]{subcaption}
\usepackage{amsmath}
\usepackage{siunitx}

\usepackage{enumitem}
\usepackage{float}
\usepackage{seqsplit}
\usepackage{framed}
\usepackage{tikz}
\usepackage{listings}
\usepackage{colortbl}
\usepackage{wrapfig}
\usepackage[most]{tcolorbox}
\usepackage{pdfpages}
\tcbuselibrary{listingsutf8}
\definecolor{mycolor}{RGB}{50,80,150}

\usepackage{ragged2e}
\usepackage{makecell}
\usepackage{adjustbox}

\usepackage[symbol]{footmisc}
\newcolumntype{Y}{>{\RaggedRight\arraybackslash}X}

\hypersetup{
    colorlinks=true,
    linkcolor=blue, 
    citecolor=blue,  
    filecolor=black,
    urlcolor=blue    
}
\definecolor{AbstractBgColor}{HTML}{F4F7FB}
\usepackage{tocloft}

\usepackage{etoolbox}
\usepackage{arydshln}
\makeatletter
\patchcmd{\@tocline}
    {\hfil}
    {\leaders\hbox{\hfil}\hfil}
    {}{}
\makeatother

\begin{document}
\sloppy
\thispagestyle{firststyle}

\begin{tcolorbox}[
    colback=AbstractBgColor, 
    colframe=AbstractBgColor, 
    arc=5pt,                  
    auto outer arc,
    boxrule=0pt,              
    left=8pt, right=8pt, 
    top=4pt, bottom=4pt,      
    parbox=false,
    width=\textwidth,
    before skip=-800pt,        
    after skip=0pt, 
    enlarge top by=-12pt
]
\begin{abstract}
\vspace{-10pt}
We introduce \ProjectName, a 35B Mixture-of-Experts Agentic Model that reaches trillion-parameter-level performance by scaling the agent horizon. We investigate agent-horizon scaling from two perspectives: scaling long-horizon trajectories and scaling heterogeneous agent abilities. To support this goal, we build a long-horizon knowledge-action infrastructure that connects external knowledge, actions, observations, and verifier outcomes, producing agentic trajectories with an average length of 45K tokens.~Based on this, we train \ProjectName~with a three-stage recipe. First, we perform full-domain supervised fine-tuning to align the base model with broad agentic behaviors. Second, we train domain-level teacher models to capture specialized expertise in each domain. Third, we propose a multi-teacher domain-routed on-policy distillation with salient vocabulary alignment to improve knowledge transfer efficiency across different domains, unifying six heterogeneous domains into one deployable student model.

\ProjectName~achieves strong and broad performance for long-horizon agent benchmarks. Compared with 1T-parameter model such as Kimi-K2.6 and DeepSeek-V4-pro, \ProjectName~achieves leading results on SEAL-0 (56.4), IFBench (80.6), HiPhO (46.4), FrontierScience-Olympiad (79.0), and MolBench-Bind (56.8), and remains highly competitive on SciCode (44.3), HLE (47.6) and BrowseComp (75.5). We hope this work provides the community with a practical path for scaling the horizon using a 35B agent that can reach or match the performance of 1T models on long-horizon tasks.
\end{abstract}
\newpage

\maketitle

\begin{center}
\href{https://github.com/InternScience/Agents-A1}{\textcolor{black}{\github~\textbf{Code}}} \qquad
    \href{https://huggingface.co/InternScience/Agents-A1}{\raisebox{-0.15\height}{\includegraphics[height=1em]{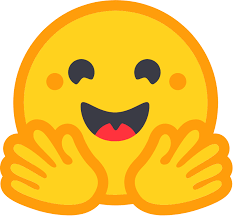}}~\textcolor{black}{\textbf{Model}}}
\end{center}
\end{tcolorbox}

\begin{figure}[H]
    \centering
    \includegraphics[width=\linewidth]{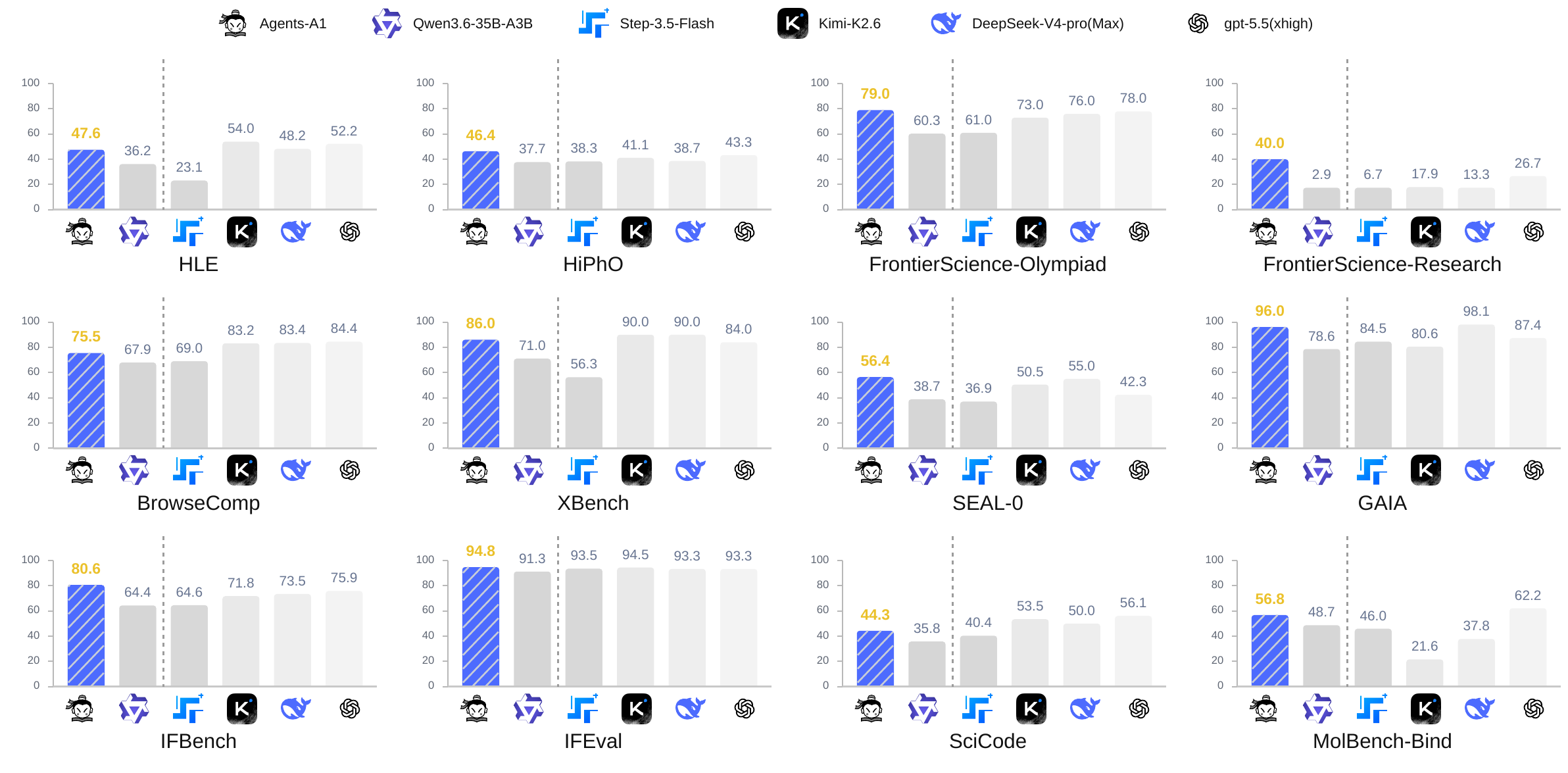}
    \caption{Benchmark performance of \ProjectName
    }
    \label{fig:benchmark_performance}
\end{figure}

\clearpage
\tableofcontents
\newpage

\section{Introduction}
\label{sec:introd}

Recent progress in LLMs~\citep{kimi2.6, GPT5.5, claude-4.6, gemini-3-pro, deepseekai2026deepseekv4, Qwen3.5} is rapidly pushing AI from passive language models toward autonomous agents that can plan, use tools, interact with environments, and improve through feedback. In real-world scenarios such as software engineering~\citep{mlebench}, scientific research~\citep{frontierscience, phan2025humanity}, and complicated decision making~\citep{browsecomp}, agents must operate over long horizons: they need to acquire information, decompose tasks, call tools, verify intermediate results, and continuously adjust their strategies. Such long-horizon settings are especially challenging since early mistakes can accumulate and strategies often need to be revised as new external information becomes available.

Existing efforts to improve long-horizon agents \cite{zeng2026glm,du2026towards,GLM5.2,team2026kimi,sun2025scaling,copet2025cwm} broadly follow two scaling routes. One route \cite{GLM5.2,team2026kimi,claude-4.6} scales parameters: frontier models commonly rely on scaling model parameters to internalize a wide range of reasoning patterns, tool-use behavior patterns, and domain knowledge. This route is effective, but it largely makes it difficult to reproduce the same agentic competence without comparable model scale, data, and training resources. The other route scales the horizon rather than enlarging parameters, which makes the intermediate decision process explicit, turning knowledge acquisition, action execution, observation interpretation, and verification into trainable supervision. However, this route also exposes two key bottlenecks.

A key bottleneck lies in the knowledge infrastructure required to support the scaling of long-horizon trajectories. Long-horizon trajectory training requires a unified environment that connects external knowledge, agentic actions, observations, and verification signals, enabling models to learn from grounded feedback rather than isolated text supervision. Without such a knowledge-tool interaction infrastructure, agents can hardly acquire the ability to plan over extended trajectories, invoke tools appropriately, verify evidence, incorporate feedback, and recover from failures in realistic settings.

Beyond infrastructure, scaling the horizon also requires integrating a broad set of heterogeneous and compositional abilities across domains. These abilities include multi-step information retrieval, tool use, executable iteration, constraint tracking, and result reflection. They often emerge unevenly across domains and interact with each other in complex ways. As a result, it is challenging to effectively combine highly different specialized abilities into one unified agentic model.

In this work, we introduce \ProjectName, a 35B MoE agentic model designed to address the key challenges mentioned above. To support agentic model training, we build a knowledge-action infrastructure that connects external knowledge, intermediate actions, observations, execution results, and verification signals, producing agentic trajectories with an average length of 45K tokens. Based on this infrastructure, we train \ProjectName~using a three-stage training recipe. First, we perform full-domain supervised fine-tuning to obtain a general agentic model with broad long-horizon abilities. Next, we train domain-level teacher models to achieve improvements in specialized domains. Finally, we propose a domain-routed on-policy distillation (OPD) with salient vocabulary alignment to unify capabilities from six heterogeneous domains into a single deployable student model.

It also aims to provide the community with a scalable technical path from model-parameter scaling to agent-horizon scaling. As shown in Fig.~\ref{fig:benchmark_performance}, \ProjectName~outperforms 1T-parameter models (Kimi-K2.6~\citep{kimi2.6} and DeepSeek-V4~\citep{deepseekai2026deepseekv4}) on SEAL-0~\citep{sealqa}, IFBench~\citep{IFBench}, HiPhO~\citep{hipho}, FrontierScience-Olympiad~\citep{frontierscience}, and MolBench-Bind~\citep{zhang2026molclaw}. It also achieves strong results on SciCode~\citep{scicode}, HLE~\citep{phan2025humanity}, and BrowseComp~\citep{browsecomp}. Our contributions include the following three aspects:
\begin{itemize}
    \item We present \ProjectName, 35B MoE model designed to scale heterogeneous agentic abilities across multiple domains. \ProjectName~can match or even outperform 1T-parameter models in long-horizon interactive agent capabilities for science and research.
    \item We build a Long-Horizon Knowledge-Action Infrastructure for multi-turn interaction with external knowledge and tools in long-horizon tasks. This infrastructure improves the agent’s ability to obtain useful information, summarize key issues, call the right tools, and execute and verify tasks.
    \item We propose a Domain-Routed On-Policy Distillation with Salient Vocabulary Alignmen to reduce conflicts caused by different reasoning patterns across domains when scaling the horizon.
\end{itemize}

\section{Knowledge-Guided General Agent Training with Specialized Teachers}
\label{sec:infra}
\subsection{Overview}

\begin{figure}[t]
    \centering
    \includegraphics[width=\linewidth]{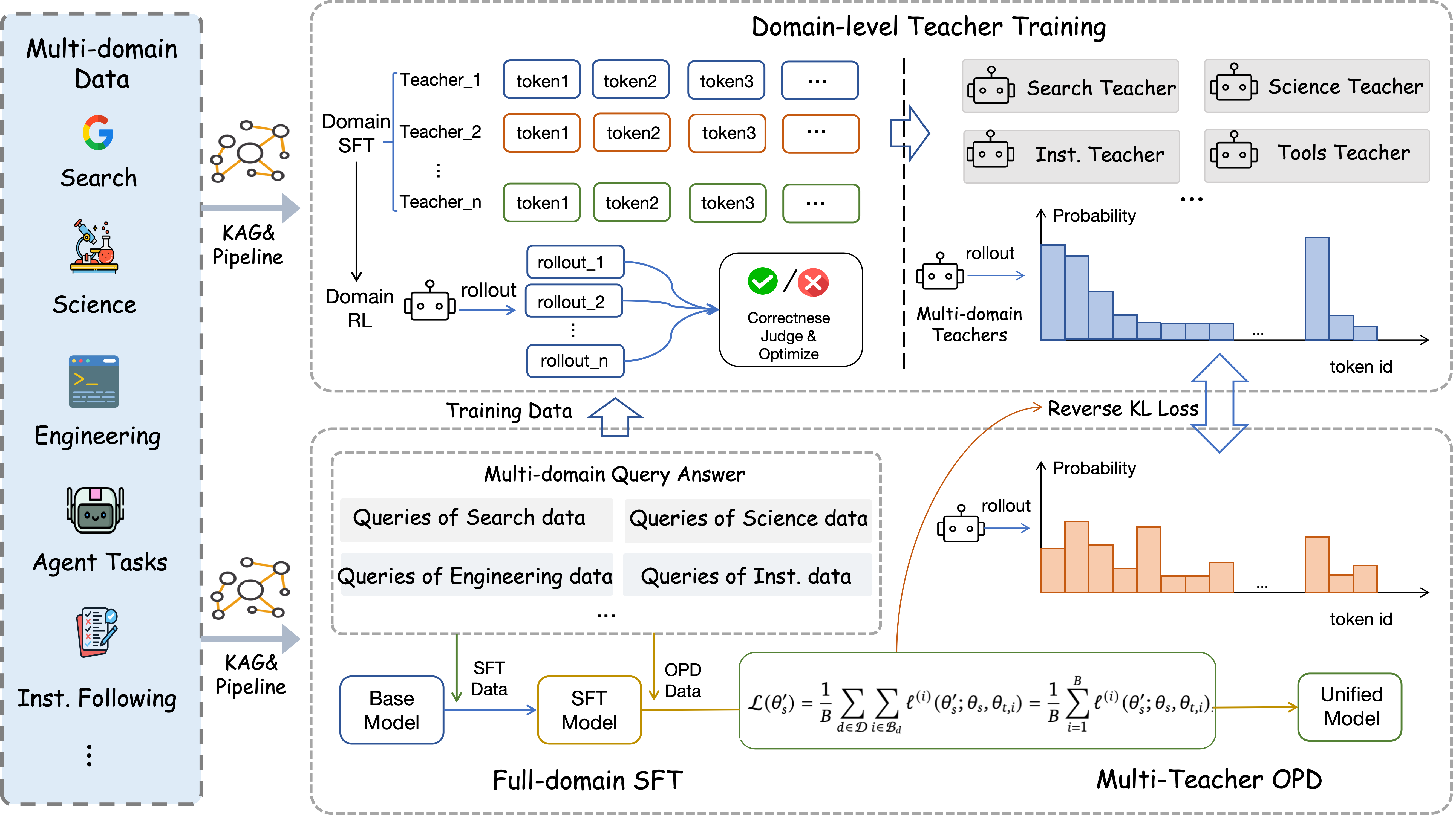}
    \caption{Overview of the three-stage training pipeline of \ProjectName. From multi-domain data to domain-specific teachers and multi-teacher on-policy distillation. First, Agents-A1 is trained with full-domain supervised fine-tuning on multi-domain long-horizon data, including search, scientific research, engineering, agentic tasks, and instruction following. Then, domain-specific teacher models are trained on each domain, and their expertise is transferred to the student model through domain-routed on-policy distillation with salient vocabulary alignment.
    }
    \label{fig:training_framework}
\end{figure}

As shown in Figure~\ref{fig:training_framework}, the training process follows a three-stage pipeline. Full-domain supervised fine-tuning is first performed to obtain a broadly capable long-horizon agent. Domain-level teachers are then trained with targeted SFT or RL, enabling each teacher to specialize in a particular capability or interaction pattern. Finally, these teachers are consolidated into a single deployable student through multi-teacher on-policy distillation.

Within this three-stage pipeline, scaling the horizon further depends on both multi-domain data infrastructure and the integration of specialized agent capabilities. For multi-domain data infrastructure, a knowledge-action graph (KAG) is constructed to retain evidence, actions, observations, failures, and verifier outcomes, providing process-level supervision beyond final answers, as described in Sec.~\ref{subsec:knowledge-action-supervision}. For integrating specialized agent capabilities, multi-teacher OPD combines domain teachers into a unified student through routed teacher guidance, salient vocabulary alignment, and domain-aware aggregation, as detailed in Sec.~\ref{subsec:on-policy-distillation}.

\subsection{Long-Horizon Knowledge-Action Infrastructure}
\label{subsec:knowledge-action-supervision}

\begin{figure}[t]
\centering
\includegraphics[width=\linewidth]{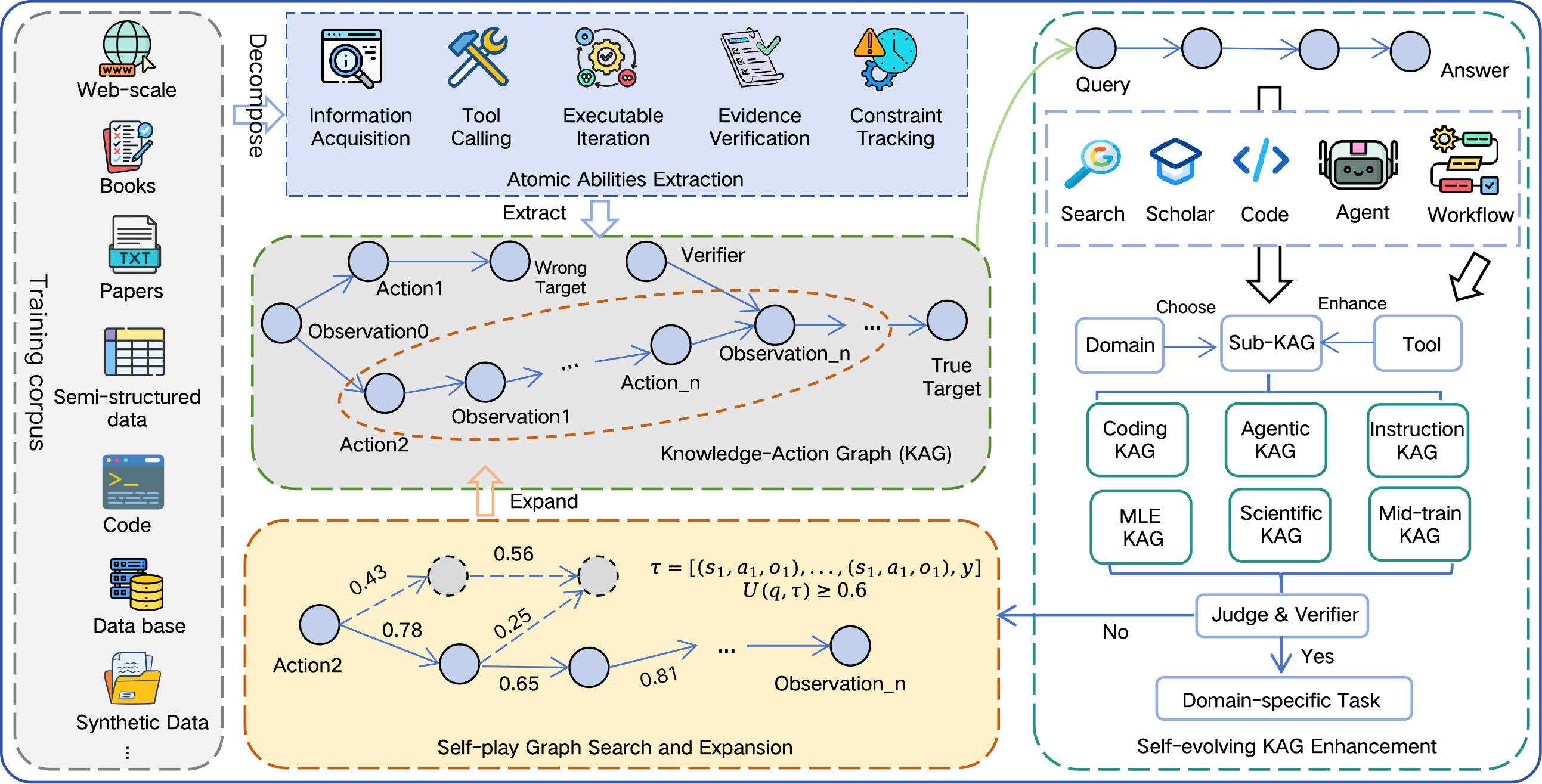}
\caption{Overview of the knowledge-action infrastructure of \ProjectName. Heterogeneous corpora are decomposed into atomic abilities and organized into a knowledge-action graph (KAG) that records evidence, actions, observations, and verifier outcomes. A tool-augmented self-play loop expands the KAG into domain-specific sub-KAGs for downstream task construction.}
\label{fig:kag_infra}
\end{figure}

As LLMs scale, training is increasingly constrained by the availability of high-density, verifiable, and evolvable supervision. Since public web corpora rarely expose provenance, action traces, tool transcripts, execution logs, and verifier outcomes, we construct a knowledge-action infrastructure that converts heterogeneous corpora into compositional, verifiable, and self-extending supervision, as shown in Figure~\ref{fig:kag_infra}.

\subsubsection{Knowledge-Action Graph Construction with Atomic Abilities}
\label{sec:domain-infrastructure}

We decompose long-horizon agentic competence into five atomic abilities: information acquisition, tool calling, executable iteration, evidence verification, and constraint tracking. To recover supervision for these abilities, we represent evidence, actions, observations, and verifier outcomes as first-class linked objects in a domain-specific knowledge-action graph (KAG). Motivated by previous work~\cite{cao2026agents}, we define the KAG as follows.

\begin{definition}[Knowledge-action graph]
Given a domain $\mathcal{B}_d$ ($d$ denotes the domain), the knowledge-action graph is a typed 4-tuple
\begin{equation}
\mathcal{G}_d = (\mathcal{C}_d,\ \mathcal{A}_d,\ \mathcal{O}_d,\ \mathcal{V}_d),
\label{eq:kag}
\end{equation}
where $\mathcal{C}_d$ is the domain \emph{corpus}, containing evidence chunks, entities, facts, constraints, and other contextual resources of the domain; $\mathcal{A}_d$ is the \emph{action space}, containing tool calls, retrieval queries, code edits and executions, reasoning steps, and other agentic operations available in $\mathcal{B}_d$; $\mathcal{O}_d$ is the \emph{observation space}, containing tool returns, retrieved evidence, execution states, and intermediate artifacts produced by executing actions on $\mathcal{C}_d$; and $\mathcal{V}_d$ is the \emph{verifier set}, containing automatic checks over correctness, evidence support, constraint satisfaction, and goal completion. The graph is populated by linked t-th action records $(s_t, a_t, o_t, v_t)$ with $s_t \subseteq \mathcal{C}_d \cup \mathcal{O}_{<t}$, $a_t \in \mathcal{A}_d$, $o_t \in \mathcal{O}_d$, and $v_t \in \mathcal{V}_d$. Edges between records encode support, dependency, production, verification, and action-transition relations.
\end{definition}

Unlike a conventional knowledge graph that mainly stores entity-relation facts, a KAG preserves the process by which an answer is acquired, tested, revised, and verified through the action records $(s_t, a_t, o_t, v_t)$. This process-level structure retains both successful and failed evidence-backed trajectories, enabling cross-step credit assignment and reproducible long-horizon supervision. 

For example, a long-horizon search task (Section~\ref{sec:search_data}) asks the agent to search an answer entity by navigating hyperlinks through a wiki corpus and web content, where $\mathcal{C}_d$ holds the pages and paragraph evidence, $\mathcal{A}_d$ is the next-hop choice, $\mathcal{O}_d$ the retrieved page text, and $\mathcal{V}_d$ checks whether the answer and its supporting path are recovered. A machine learning engineering task (Section~\ref{sec:mle_data_pipeline}) asks the agent to optimize a Kaggle-style submission by iteratively writing, patching, and executing code over a tree of candidate solution, where $\mathcal{C}_d$ holds the competition specification, dataset, and execution environment, $\mathcal{A}_d$ is a code edit, run, or commit action, $\mathcal{O}_d$ is the resulting log, metric, or submission artifact, and $\mathcal{V}_d$ checks grader scores and submission validity.

\subsubsection{Self-play Graph Search and Expansion}
\label{sec:game-generation}

In practice, the reasoning data generated in a single pass is not necessarily of high quality; it requires multiple iterations and validation to produce.
To optimize and improve the quality of $\mathcal{G}_d$ in each domain, we expand $\mathcal{G}_d$ through a proposer--solver--verifier game: $\pi_{\text{P}}$ samples graph regions to propose constrained tasks, $\pi_{\text{S}}$ solves them with retrieval and tools, and $\pi_{\text{V}}$ verifies answers, evidence, execution results, trajectories, and shortcut risks. Specifically, each generated task is represented as
\begin{equation}
x = (q,\ d,\ \tau,\ y^\star,\ \mathcal{E}_q,\ \mathcal{V}_q),
\qquad
\mathcal{E}_q \subseteq \mathcal{C}_d,\quad \mathcal{V}_q \subseteq \mathcal{V}_d,
\label{eq:task}
\end{equation}
where $q$ is the instruction, $d$ the domain label, $y^\star$ the target answer, $\mathcal{E}_q$ the supporting evidence required for the task, $\mathcal{V}_q$ the verifier subset applicable to the task, and $\tau$ the resulting trajectory
\begin{equation}
\tau = \big[(s_1, a_1, o_1, v_1),\ \ldots,\ (s_T, a_T, o_T, v_T),\ y\big],
\qquad
a_t \in \mathcal{A}_d,\ o_t \in \mathcal{O}_d,\ v_t \in \mathcal{V}_q \cup \{\bot\},
\label{eq:trajectory}
\end{equation}
following the action-record form of Eq.~\ref{eq:kag}, with $v_t = \bot$ when no step-level verifier fires.
A candidate $x$ is accepted by $\pi_{\text{V}}$ only if it is (i) verifiable against some $v \in \mathcal{V}_q$, (ii) valid, i.e., $\tau$ reaches an answer that $v$ accepts, (iii) process-informative, i.e., $\tau$ exercises meaningful intermediate decisions rather than collapsing to a one-shot lookup, (iv) evidence-covering, i.e., the required evidence in $\mathcal{E}_q$ is actually consulted along $\tau$, and (v) unambiguously specified, with no shortcut solution.

In this way, solver and verifier feedback is written back to $\mathcal{G}_d$ as linked states, actions, observations, evidence, artifacts, and verifier outcomes. Given a query-answer pair and an initial KAG, the framework invokes domain tools, including search, scholar, code execution, agent modules, and workflow planners, to enhance the graph. The enhanced graph is specialized into sub-KAGs for coding, agentic reasoning, instruction following, MLE, and scientific reasoning. A judge-and-verifier module accepts qualified sub-KAGs for downstream task-pipeline construction and routes failed ones back to self-play expansion.

\subsection{Domain-Routed On-Policy Distillation with Salient Vocabulary Alignment}
\label{subsec:on-policy-distillation}

Domain-specific teachers provide specialized policies, but deployment requires a single general policy model. Existing sampled-token OPD~\citep{lu2025onpolicydistillation} is efficient because it scores only realized rollout tokens, but this single-token approximation leaves nearby high-probability alternatives unconstrained, a source of unstable or imbalanced guidance noted in recent analyses~\citep{fu2026revisiting,deepseekai2026deepseekv4}.

We therefore use a domain-routed multi-teacher OPD framework with salient vocabulary alignment (SVA). For each prompt-domain pair $(x_i,d_i)$, a frozen rollout student samples $y_i\sim\pi_{\theta_s}(\cdot\mid x_i)$, while the optimized student $\theta_s'$ is supervised by the routed teacher $\theta_{t,i}\triangleq\theta_t^{d_i}$. SVA replaces the sampled-token surrogate by aligning the student and routed teacher on a compact teacher-supported local vocabulary. Losses are computed only on trainable generated tokens $R_i$, with tool outputs and user turns masked out. To handle cross-domain heterogeneity, we aggregate SVA losses with a domain-normalized objective, averaging within each active domain and then across active domains, so that frequent or high-loss domains do not dominate the student update.

\subsubsection{Salient Vocabulary Alignment}

At position $t$, SVA evaluates the current student and the routed teacher on the same student-generated prefix $(x_i,y_{i,<t})$. Let
\[
p_{s'}(u)=\pi_{\theta_s'}(u\mid x_i,y_{i,<t}),
\qquad
p_{t,i}(u)=\pi_{\theta_{t,i}}(u\mid x_i,y_{i,<t}),
\]
and let $\mathcal{S}_{i,t}^{(k)}$ be the set of top-$k$ valid tokens under the routed teacher distribution. We renormalize both distributions on this teacher-selected support
\[
\bar p_{s'}(u)
=
\frac{p_{s'}(u)}
{\sum_{v\in\mathcal{S}_{i,t}^{(k)}}p_{s'}(v)},
\qquad
\bar p_{t,i}(u)
=
\frac{p_{t,i}(u)}
{\sum_{v\in\mathcal{S}_{i,t}^{(k)}}p_{t,i}(v)},
\quad
u\in\mathcal{S}_{i,t}^{(k)}.
\]

The per-sample SVA objective is the truncated reverse KL over this salient support, averaged over trainable model-generated positions
\begin{equation}
\ell_{\mathrm{SVA}}^{(i)}(\theta_s';\theta_{t,i})
=
\frac{1}{|R_i|}
\sum_{t\in R_i}
\sum_{u\in \mathcal{S}_{i,t}^{(k)}}
\bar p_{s'}(u)
\log
\frac{\bar p_{s'}(u)}{\bar p_{t,i}(u)} .
\end{equation}

Because the support is teacher-selected, SVA does not directly constrain student mass outside $\mathcal{S}_{i,t}^{(k)}$. We therefore monitor the student-side coverage
\begin{equation}
\rho(i,t)
=
\sum_{u\in \mathcal{S}_{i,t}^{(k)}} p_{s'}(u),
\end{equation}
where higher coverage indicates a closer approximation to full-vocabulary alignment.

\subsubsection{Domain-routed Normalized Objective}

In the multi-domain setting, different domains may induce heterogeneous gradients because they emphasize different capabilities, such as response style, reasoning pattern, evidence acquisition, tool use, or external interaction. We use hard domain routing, where each sample is supervised only by the teacher trained for its domain,
\[
\theta_{t,i} = \theta_t^{d_i},
\]
rather than by a soft mixture of teachers. This preserves domain-specific teacher preferences and avoids mixing incompatible teacher signals at the token level.

For a mini-batch $\mathcal{B}$, let $\mathcal{B}_d$ denote the subset of samples from domain $d$, and let
\[
\mathcal{D}_{\mathcal{B}}
=
\{d\in\mathcal{D}\mid |\mathcal{B}_d|>0\},
\]
be the set of active domains. Since each per-sample SVA loss is already averaged over its own trainable response positions, longer trajectories do not dominate merely by contributing more tokens. To further prevent high-frequency domains from dominating the update, we average losses within each active domain and then average over active domains:
\begin{equation}
\mathcal{L}_{\mathrm{MT\text{-}SVA}}(\theta_s')
=
\frac{1}{|\mathcal{D}_{\mathcal{B}}|}
\sum_{d\in \mathcal{D}_{\mathcal{B}}}
\frac{1}{|\mathcal{B}_d|}
\sum_{i\in \mathcal{B}_d}
\ell_{\mathrm{SVA}}^{(i)}(\theta_s';\theta_{t,i}).
\end{equation}
This domain-normalized objective gives each active domain comparable influence while allowing each sample to receive supervision from its corresponding domain-specific teacher. Domain labels are also used to instantiate domain-specific rollout protocols, such as finalization rules, simulated user turns, or environment-specific tool execution. After rollouts are generated, all trainable positions are optimized using the same SVA objective under their routed teachers.

\section{Multi-domain Data Pipeline}
\subsection{Long-horizon Search}
\label{sec:search_data}
We instantiate the knowledge-action graph (KAG) in the search domain by constructing search question data over a large wiki database. Here $\mathcal{C}_d$ is a corpus of wiki pages and paragraph-level evidence, $\mathcal{A}_d$ consists of entity transitions induced by hyperlinks and controlled graph walks, $\mathcal{O}_d$ contains the retrieved page text and connecting evidence, and $\mathcal{V}_d$ checks whether the answer entity and its supporting path are recoverable from the provided context. This instantiation targets the information acquisition and evidence verification abilities defined in Section~\ref{sec:domain-infrastructure}. The pipeline first converts wiki entries into a directed corpus graph, where each entry is an entity node and inter-entry references define edges. For each node, the pipeline retrieves page text, paragraph structure, outgoing anchors, canonical titles, and optional structural statistics such as in-degree, out-degree, and text length. Non-content tail sections and anchors from list-style paragraphs are filtered to reduce noisy transitions, so the resulting graph can provide both the search space and the provenance records required by $\mathcal{G}_d$.

\textbf{Relation-chain generation.} Relation chains are generated through controlled random walks over the corpus graph and the corresponding domain KAG. Candidate next nodes are selected from outgoing anchors after removing visited entities, near-duplicate title variants, disambiguation pages, pages without valid text, and pages with insufficient outgoing links. Degree and text-length constraints further control node quality and graph diversity. An LLM selector chooses one entity from the remaining candidates, balancing local path coherence with optional cross-domain topic shifts. If a walk reaches a dead end, a short auxiliary walk recovers a qualified continuation node. Each accepted transition is stored as an action-like record linking the current entity state, the selected hyperlink action, the observed target page, and the paragraph evidence that justifies the transition.

\textbf{Question-answer pair generation.} Each completed chain is serialized with its ordered entity list, node texts, and paragraph-level evidence for adjacent entities. The generator treats the final entity as the answer, rewrites the chain into a coherent natural-language question, masks proper nouns and other identifying information, and requires the solver to recover the masked answer entity. A verifier then checks the answer identity and evidence support against the serialized chain. Each chain, therefore, yields a verifiable search-domain training instance that requires following indirect long-context evidence rather than relying on direct name matching, while preserving the provenance, action path, observation text, and verifier target needed by the KAG.

\textbf{Trajectory collection and quality control.} Search trajectories are collected by allowing strong models to execute deep-research tasks with the \texttt{search}, \texttt{read\_page}, and \texttt{code} tools in the real Internet environment. The \texttt{search} tool queries a commercial search engine and returns the top results per query. The \texttt{read\_page} tool extracts web-page content and uses an LLM to summarize the extracted information before it is returned to the agent context. The \texttt{code} tool allows the model to write and execute Python code in a sandboxed environment, supporting intermediate computation and data processing during the search process. The maximum context window is set to 256K tokens, and no additional constraint is imposed on the number of tool calls within a single turn. During post-processing, we remove trajectories with wrong answers, overly short interaction histories, or obvious guessing behavior. For trajectories that reach the maximum turn limit, we roll back to the previous turn and use an explicit user prompt to force the model to produce an answer. The retained trajectories provide long-horizon supervision for search behavior that combines query formulation, page reading, evidence integration, and answer verification under realistic web observations.


\subsection{Machine Learning Engineering}
\label{sec:mle_data_pipeline}
Following the knowledge-action infrastructure in Section~\ref{subsec:knowledge-action-supervision}, we instantiate machine learning engineering (MLE) as a KAG over executable solution search. Here $\mathcal{C}_d$ contains competition specifications, datasets, and execution environments; $\mathcal{A}_d$ captures solution edits, experiment runs, tree navigation, node invalidation, and answer commitment; $\mathcal{O}_d$ records logs, metrics, artifacts, and generated submissions; and $\mathcal{V}_d$ provides grader scores, submission-format checks, and metric-reliability checks. Because a candidate solution is executable code whose quality is known only after grading, each trajectory becomes a verifier-guided expansion of the solution graph.

\textbf{Task sources.}
We assemble gradeable competitions from MLE-Dojo~\citep{mledojo} and ended Kaggle competitions processed by our automated framework. MLE-Dojo provides curated Kaggle-style tasks with held-out answers and local graders across tabular, vision, NLP, audio, and time-series settings. For ended competitions, our framework curates the task description and leaderboard information, re-splits public data into fresh train/test partitions, constructs private answers, and synthesizes a local evaluator with submission validation. This produces a diverse and refreshable pool of optimization tasks while reducing overfitting to any static benchmark distribution.

\textbf{Agentic harness.}
Trajectories are generated in an agentic harness that grows a tree of executable solution nodes, following prior solution-search workflows such as MLEvolve~\citep{mlevolve}. Writing a full script opens a new root, patching a node spawns a child, and executing a node attaches observations such as logs, exceptions, metrics, artifacts, and submission validity. This tree forms the executable core of the MLE-domain KAG: branches encode alternative attempts, verifier outcomes guide expansion, and failed or invalidated nodes provide negative evidence for later decisions.

The harness exposes this process through a compact tool interface, summarized in Table~\ref{tab:mle_tools}, covering code authoring, execution, tree navigation, answer management, persistent memory, and delegated analysis. For long runs, an isolated \texttt{analyze} sub-agent investigates data or results and returns a report, while context compaction summarizes earlier steps into a digest.

\textbf{Trajectory collection and quality control.}
We collect teacher trajectories over the task pool with multiple seeds and prompt variants, replay them with the local evaluator, and retain runs that produce valid and competitive committed submissions. After trimming regressive or degenerate segments, we deduplicate the remaining runs and serialize them into a unified message schema with loss masks for teacher-generated content. The serialization preserves node relations, execution observations, verifier outcomes, and committed-answer history, making the solution-search structure recoverable for training.

\begin{table}[ht]
\centering
\caption{\textbf{Our designed tool interface used by \ProjectName, according to the MLE agentic harness.} The tools define a compact action space through which the model performs code authoring, execution, search-tree management, persistent memory, and delegated analysis.}
\label{tab:mle_tools}
\begin{tabular}{@{}l p{0.66\linewidth}@{}}
\toprule
\textbf{Tool} & \textbf{Function} \\
\midrule
\multicolumn{2}{@{}l}{\textit{Code authoring \& execution}}\\
\texttt{write\_full\_code} & Author a complete training script from scratch; opens a new root node (a fresh line of attack). \\
\texttt{patch\_code} & Apply a localized edit to a node's code; spawns a child node, preserving tree history for incremental refinement. \\
\texttt{execute\_code} & Run a node, capture stdout and exceptions, extract its validation metric, and check the emitted submission for validity. \\
\texttt{execute\_bash} & Run a guarded shell command for environment setup and inspection (installs, GPU checks, file operations). \\
\addlinespace
\multicolumn{2}{@{}l}{\textit{Search-tree navigation \& answer management}}\\
\texttt{list\_nodes} & Survey the solution tree: the selected answer, the recent answer trail, invalidated history, and a metric-ranked listing. \\
\texttt{select\_node} & Inspect one node in full (code, plan, output, metric, parent chain) before revisiting or branching from it. \\
\texttt{invalidate\_node} & Exclude a node whose metric is untrustworthy (leakage, overfitting) from ranking and submission. \\
\texttt{update\_answer} & Commit a node as the current submission candidate, written to the canonical path the grader reads. \\
\texttt{get\_current\_answer} & Report the node currently committed as the answer. \\
\addlinespace
\multicolumn{2}{@{}l}{\textit{Persistent memory}}\\
\texttt{write\_notes} / \texttt{read\_notes} & Append to / re-read a notebook that survives context compaction (decisions, failed strategies and why, hypotheses). \\
\addlinespace
\multicolumn{2}{@{}l}{\textit{Sub-agent}}\\
\texttt{analyze} & Spawn an isolated analysis sub-agent that explores data and results in its own context window and returns a single structured report. \\
\bottomrule
\end{tabular}
\end{table}


\subsection{Scientific Reasoning and Research}
\label{sec:science-data-pipeline}
Following the knowledge-action infrastructure in Section~\ref{subsec:knowledge-action-supervision}, we instantiate scientific reasoning as a domain-specific KAG over scientific problem-solving processes. Formally, given the scientific knowledge-action graph $\mathcal{G}_d = (\mathcal{C}_d, \mathcal{A}_d, \mathcal{O}_d, \mathcal{V}_d)$,  $\mathcal{C}_d$ contains scientific problem components including problem statements, topics, keywords, and solution structures; $\mathcal{A}_d$ consists of reasoning actions such as decomposition, transformation, retrieval, and computation; $\mathcal{O}_d$ captures intermediate reasoning states, external evidence, and execution outputs; and $\mathcal{V}_d$ contains verification signals over correctness, consistency, and scientific validity.

\textbf{Problem Construction.} We first collect a large-scale pool of scientific problems spanning fundamental domains such as mathematics, physics, and related areas. Based on this seed pool, we construct a knowledge-action graph, where each seed is represented as interconnected nodes encoding its problem statement, keywords, and solution components, forming an initial knowledge-aligned subgraph. Built upon this graph, we perform self-evolving scientific KAG enhancement introduced in Section \ref{subsec:knowledge-action-supervision} via self-play graph search and expansion, where local subgraphs are sampled and systematically rewritten to improve scientific problems along two complementary directions: (i) \emph{harder reasoning variants}, which increase required domain knowledge depth, introduce complex symbolic structures, and extend multi-step derivations; and (ii) \emph{interaction-enriched variants}, which inject cross-domain knowledge, incorporate concepts requiring external retrieval, and increase computational demands such as code-based numerical execution. For problems that already exhibit strong reasoning ability or strong interaction requirements, we retain their original form without modification. Through this graph-driven expansion process, we construct an enhanced scientific problem corpus of approximately 15K instances, characterized by higher difficulty and stronger interaction, thereby providing a data foundation for long-horizon scientific reasoning.

\textbf{Trajectory Generation.} Based on the enhanced scientific problem corpus, we construct both no-tool and tool-augmented reasoning trajectories as training data for long-horizon scientific reasoning. For no-tool reasoning, we distill high-quality pure reasoning trajectories from a strong model, consisting of multi-step derivations, symbolic transformations, and final answers, and retain only those verified as correct to improve pure reasoning capability. For tool-augmented reasoning, we build an interactive reasoning framework equipped with four external tools: \texttt{search}, \texttt{visit}, \texttt{code}, and \texttt{scholar}. The \texttt{search} tool retrieves relevant web information for factual grounding; the \texttt{visit} tool enables inspection of specific web pages or documents; the \texttt{code} tool supports numerical computation, symbolic calculation, equation solving, and simulation; and the \texttt{scholar} tool provides access to academic literature. Using this framework, we distill tool-augmented trajectories from a strong model, recording intermediate reasoning steps, tool calls, observations, computations, and final answers, and filter them by final-answer correctness to ensure high-quality interaction data. Overall, no-tool and tool-augmented trajectories provide complementary supervision for strengthening reasoning and interaction capabilities in long-horizon scientific problem solving.

\subsection{Instruction Following}
\label{sec:if_data}

Following the knowledge-action infrastructure in Section~\ref{subsec:knowledge-action-supervision}, we instantiate instruction following as a KAG-style supervision problem over constraints, long-context evidence, and locally introduced rules. Here $\mathcal{C}_d$ contains user instructions, verifiable constraints, long-document chunks, document-level entities and facts, injected in-context rules, distractors, and answer candidates; $\mathcal{A}_d$ consists of KAG-level operations such as constraint parsing, evidence selection, local-rule application, distractor rejection, and final answer formatting; $\mathcal{O}_d$ contains selected evidence, parsed constraint states, injected rule states, and candidate answer states; and $\mathcal{V}_d$ contains automatic validators for constraint satisfaction, answer matching, evidence dependence, and consistency with injected in-context information. This instantiation targets constraint tracking, evidence verification, and long-context adaptation.

\textbf{Task Construction.}
We construct the instruction-following corpus from two complementary sources with the corresponding domain KAG. The first source is a high-quality subset of 13K multi-constraint instruction samples from NVIDIA's Nemotron-RL instruction-following dataset~\citep{nemo-gym}. These examples are built from prompts in WildChat-1M~\citep{zhao2024wildchat} and instructions from the Open-Instruct codebase, and cover automatically verifiable constraints such as length, format, keywords, language, punctuation, and paragraph structure. In the KAG view, each constraint is represented as a condition node linked to the required output behavior and its validator, so the resulting data directly supervises whether the model can parse and satisfy explicit user requirements.

The second source is a self-built long-context learning pipeline that produces 10K verified long-context QA instances. We first structurally parse long documents and extract entities, attributes, relations, numerical values, and other salient facts to construct document-level evidence graphs compatible with the KAG representation. Based on these graphs, we synthesize multi-hop QA tasks that require combining evidence across multiple factual nodes. We then inject local in-context rules or distractors, such as temporary protocols that override document rules, misleading framing statements, or constraints that conflict with prior knowledge. These injected signals are linked to the relevant evidence nodes, rule nodes, and verifier checks. Candidate tasks are converted into a unified multiple-choice QA format and filtered by automatic validation to ensure that the correct answer depends on both the long-context evidence chain and the injected in-context information. This yields verifiable supervision for locating dispersed evidence, applying local rules, and resisting unsupported distractors.

\subsection{Tool Calling}
\label{sec:tool-calling}

Following the knowledge-action infrastructure in Section~\ref{subsec:knowledge-action-supervision}, we instantiate tool calling as a KAG over executable interactions. Here $\mathcal{C}_d$ contains tool schemas, environment states, optional simulated user profiles, and task resources; $\mathcal{A}_d$ consists of schema-grounded calls and clarification actions; $\mathcal{O}_d$ contains tool returns, state updates, errors, and user feedback; and $\mathcal{V}_d$ contains available schema, state, grounding, and goal-completion checks. Each task can induce multiple candidate long-horizon state--action--observation chains whose later decisions depend on earlier observations.

\textbf{Task Construction.} The task pool and tool interface are constructed jointly through tool extraction, tool-interaction graph construction, graph-compositional task synthesis, and solvability assessment. We extract candidate interfaces from scientific, web, repository, database, and locally simulated tool-usage settings~\citep{barres2025tau2,he2025vitabench}. The graph connects tools, resource/state types, observation fields, and verifier targets, with edges denoting executable dependencies such as schema compatibility, precondition--effect relations, shared resources, state transitions, or support for a completion check. Task synthesis is formulated as constrained graph search over this dependency graph: the generator selects connected tool subgraphs or dependency paths anchored by target states, answers, or verifier conditions, and then renders them into user instructions. This avoids arbitrary tool concatenation: later calls must rely on objects, constraints, or observations produced earlier. Candidate tasks are screened for argument availability, initial-state reachability, sandbox executability, and verifier coverage. Underspecified, prompt-only, schema-incompatible, or unverifiable tasks are revised or discarded. Each retained task stores the user goal, selected tool subgraph, initial state, hidden support path, and completion criterion or verifier.

\textbf{Trajectory Generation.} Trajectory generation is performed in a Tool Sandbox, which exposes the selected tool subgraph and maintains the evolving environment state. Solver backends explore the candidate space over multiple turns by choosing tools, binding arguments, interpreting observations, and deciding whether to continue, ask for clarification, or produce the final response. When preferences, missing constraints, or task-bounded clarifications are required, a simulated user participates in the loop. For the same user goal and verifier target, including the same question--answer pair in QA-style tasks, we generate multiple candidate trajectories through different tool choices, actions, or clarification turns. This process is treated as verifier-guided graph search over the trajectory space. Available verifier or judge modules score or rank these candidates by call format, schema typing, state consistency, observation grounding, constraint satisfaction, and goal completion. Invalid schemas, unsupported arguments, unrecovered failures, or ungrounded responses are rejected or routed to repair and resampling. Accepted runs are kept as KAG-aligned message trajectories containing assistant turns, tool calls, observations, user feedback, state changes, and verifier outcomes.
\section{Three-stage Training Recipe}

\subsection{Full-domain Supervised Fine-Tuning}
\label{sec:sft}

We start from Qwen3.5-35B-A3B~\citep{Qwen3.5} and perform supervised fine-tuning (SFT). The goal is to make the model follow the desired behavior and improve its ability to follow instructions. SFT helps bridge the gap between atomic ability learned during mid-training and the ability to give helpful, clear, and context-aware responses to user instructions.

\textbf{Data Composition.}
Our SFT dataset comprises a diverse mixture of high-quality long-horizon trajectories spanning multiple domains and task categories. The data sources include:
\begin{itemize}
    \item \textbf{Deep research}: Surveys, deep research topics, and puzzles that require the model to search the internet for relevant information and synthesize answers.
    \item \textbf{Coding and engineering}: Programming tasks spanning multiple languages and frameworks, with step-by-step reasoning and code generation. This category also includes machine learning engineering data, where the model develops machine learning methods such as designing model architectures, implementing training pipelines, and tuning hyperparameters.
    \item \textbf{Scientific problem-solving}: Chain-of-thought solutions to scientific and mathematical problems, including the use of scientific computation tools.
    \item \textbf{Instruction following}: Tasks designed to enhance the model's ability to follow fine-grained instructions in generative settings under strict constraints.
    \item \textbf{General agentic tasks}: Multi-turn dialogues involving planning, decision-making, and general tool-use capabilities.
\end{itemize}

In total, the SFT dataset contains approximately $100$K trajectories with an overall average length of $45$K tokens. Table~\ref{tab:sft_data_stats} summarizes the average token length per domain. Notably, the majority of our SFT data consists of long-horizon trajectories, which enhance the model's long-horizon thinking capabilities under different extended contexts. All data undergo rigorous quality filtering, deduplication, and human review to ensure accuracy and consistency.

\begin{table}[ht]
\centering
\caption{Average token length per data source domain in the SFT dataset.}
\label{tab:sft_data_stats}
\begin{tabular}{lc}
\toprule
\textbf{Data Source} & \textbf{Avg.\ Token Length} \\
\midrule
Deep research & 44K \\
Coding and engineering & 48K \\
Scientific reasoning and problem-solving & 37K \\
Instruction following & 3K \\
General agentic tasks & 39K \\
\midrule
\textbf{Overall} & \textbf{45K} \\
\bottomrule
\end{tabular}
\end{table}

\textbf{Training Details.} We fine-tune from Qwen3.5-35B-A3B~\citep{Qwen3.5} using a standard cross-entropy loss computed only on the response tokens, while the instruction tokens are masked from the loss computation. This ensures the model learns to generate responses conditioned on given instructions without overfitting to prompt patterns. Key hyperparameters are listed in Table~\ref{tab:sft_hyperparams}.

\begin{table}[ht]
\centering
\caption{Hyperparameters for supervised fine-tuning.}
\label{tab:sft_hyperparams}
\begin{tabular}{ll}
\toprule
\textbf{Hyperparameter} & \textbf{Value} \\
\midrule
Learning rate & $1 \times 10^{-5}$ \\
Learning rate schedule & Cosine with warmup \\
Warmup ratio & $0.05$ \\
Batch size & $16$ \\
Epochs & $1$ \\
Max sequence length & $131{,}072$ \\
Optimizer & AdamW \\
Weight decay & $0.1$ \\
\bottomrule
\end{tabular}
\end{table}

To improve training throughput, we adopt a sample packing strategy that concatenates multiple short examples into a single training sequence up to the maximum context length. Attention masks are applied to prevent cross-contamination between packed samples. This reduces padding overhead and leads to significant improvements in GPU utilization. 

\subsection{Domain-level Teacher Training}
\label{sec:domain-level}

\subsubsection{Reinforcement Learning on Search Tasks}
To construct the teacher model of agentic searching, we adopt a two-stage training pipeline. In the first stage, we perform supervised fine-tuning (SFT) on the base model using only the search trajectories collected in Section~\ref{sec:search_data}. These trajectories demonstrate how to decompose complex questions into sub-queries, invoke web search and page reading tools at appropriate points, and synthesize retrieved information into a coherent answer. The SFT stage equips the model with basic tool-use capabilities and multi-turn agentic search patterns. In the second stage, we apply RL on top of the SFT model to further improve the model's ability to leverage external search tools for complex multi-hop reasoning.

\textbf{RL Algorithm.} We adopt GRPO~\citep{shao2024deepseekmath} as our RL algorithm. The training objective combines a clipped policy loss, a KL divergence penalty to prevent the policy from deviating too far from the reference model, and an entropy regularization term to encourage exploration. We use rollout log-probabilities for accurate advantage computation.

\textbf{Agentic Tools.} We equip the model with three tools during RL training:
\begin{itemize}
    \item \textbf{Web Search}: Returns Google search results given a query, allowing the model to discover relevant web pages.
    \item \textbf{Read Page}: Given a URL and a query, this tool uses a summarization model to extract and return information from the page that is relevant to the query.
    \item \textbf{Code}: Enables the model to write and execute Python code in a sandbox environment, supporting computation and file processing.
\end{itemize}

\textbf{Training Data.} The training dataset consists of approximately 2,000 carefully selected multi-hop reasoning questions that require web search capabilities. Each sample contains a user query and a ground-truth answer. To construct the dataset, we use the SFT model equipped with the above tools to attempt each problem with 5 retries. We then select the questions for which the model produces both correct and incorrect trajectories, filtering out questions that are either too easy (always correct) or too hard (always incorrect) for the model. This selection strategy ensures that the RL training signal is maximally informative.

\textbf{Rollout and Reward Design.}
During each rollout step, the model generates 8 rollouts per prompt. Each response is a multi-turn agentic trajectory in which the model iteratively invokes tools and reasons over the retrieved information. The final reward combines three components:

\textit{(1) Correctness reward.} We employ an LLM judge to evaluate whether the model's final answer is correct. The judge accepts equivalent numerical formats, semantic paraphrases, and answers that contain the correct information as a clear sub-phrase, while rejecting contradictory or evasive responses.

\textit{(2) Search behavior penalties.} We introduce two penalties to encourage efficient and non-redundant search behavior.
\begin{itemize}
    \item \textbf{Efficiency penalty}: We allow the model to search freely within the first $K$ rounds without any penalty. Beyond $K$ rounds, a penalty that increases linearly with the number of additional rounds is applied. This encourages the model to stop searching promptly once sufficient information has been gathered.
    \item \textbf{Repetition penalty}: If a \texttt{google\_search} query or a \texttt{read\_page} URL has already appeared within a recent sliding window, each repeated occurrence incurs a small penalty, capped at a maximum per trajectory. This discourages redundant searches and repetitive page reads.
\end{itemize}

\textit{(3) Format calibration reward.} We additionally reward the model for producing answers that conform to the expected output format, ensuring that the final response is well-structured and can be reliably parsed for evaluation.

\textbf{Training Configuration.} We fine-tune the SFT model using RL with a constant learning rate of $1 \times 10^{-6}$. At each rollout step, we sample 32 questions and generate 8 candidate responses per question, yielding a global batch size of 256 for each gradient update. The GRPO objective uses a clip range of $[0.2, 0.28]$, a KL divergence penalty coefficient of $0.001$, and an entropy regularization coefficient of $0.0001$. The maximum response length is set to 4,096 tokens and the maximum sequence length is 131,072. Each rollout has 300 tool calls at most. The rollout temperature is set to 1.0.


\subsubsection{Science-enhanced Supervised Fine-Tuning}
In this part, we introduce how to train a science teacher model that preserve both deep problem-solving ability and reliable interactive behavior in scientific scenarios. Because frontier scientific tasks often combine long derivations, specialized knowledge, numerical computation, and symbolic manipulation, a science teacher requires to maintain both strong intrinsic reasoning and extrinsic interaction abilities. It should reason through difficult scientific and professional problems, decide when internal reasoning is insufficient, invoke external tools when factual grounding or exact computation is needed, and synthesize the resulting evidence into a coherent answer. Thus, we design a two-stage SFT pipeline. The first stage focuses on boosting the \textbf{reasoning ability} while the second stage further strengths the \textbf{interaction ability}. The details are as following. 

\textbf{Training Data.}
We initialize from the general-domain model Qwen3.5-35B-A3B~\citep{Qwen3.5} and continue training it with the scientific data constructed in Section~\ref{sec:science-data-pipeline}. The supervised corpus contains two complementary types of targets. The first type consists of high-quality reasoning traces produced by strong non-tool reasoning models over curated scientific QA problems. These traces emphasize intrinsic scientific reasoning ability, including problem decomposition, physical assumption identification, derivation of intermediate quantities, unit consistency, and final-answer verification. The second type consists of filtered tool-augmented reasoning trajectories equipped with \texttt{search}, \texttt{visit}, \texttt{code}, and \texttt{scholar}. These trajectories expose the model to grounded interaction with external objects, including retrieving background facts, inspecting source documents, executing computations, checking intermediate results, and using the observations to revise the solution path.

\textbf{Two-stage SFT.} The science teacher is trained through a two-stage SFT pipeline. Both stages use the standard response-token supervised objective.
\begin{itemize}
    \item \textbf{Stage 1: Reasoning-Enhanced SFT.} 
     we focus on substantially strengthening the model's intrinsic reasoning depth. We fine-tune the model on high-quality non-tool scientific reasoning trajectories, which encourages the model to form complete causal and mathematical chains before producing the final response. This stage improves the teacher's ability to solve problems through self-contained derivation rather than relying prematurely on extrinsic tools.

    \item \textbf{Stage 2: Tool-Augmented SFT.}
    we continue from the strong-reasoning checkpoint and perform tool-enhanced SFT on the filtered multi-subject science trajectories. Compared with the first-stage data, these samples require the model to coordinate reasoning with explicit tool interactions. The training objective encourages the teacher to recognize when a problem requires external support, choose the appropriate tool, formulate precise tool inputs, interpret returned observations, validate numerical or symbolic intermediate states, and fold retrieved or computed evidence back into a logically connected solution. We extend the training rounds in this stage to make tool-use behavior stable under long scientific interactions, while retaining the reasoning style acquired in the first stage.
\end{itemize}



\subsubsection{Reinforcement Learning on Instruction Following}

We hypothesize that stronger instruction-following ability can facilitate effective in-context learning (ICL), especially in long-context settings where the model must understand task instructions, respect output constraints, and identify relevant evidence from lengthy inputs. To this end, we adopt a two-stage reinforcement learning pipeline to optimize our SFT model. The first stage focuses on fine-grained instruction following, while the second stage further improves the model's ICL ability.

\textbf{Training Data.}
We use the instruction-following and long-context learning data constructed in Sec.~\ref{sec:if_data}. The first RL stage uses the verifiable instruction-following subset to optimize fine-grained constraint satisfaction, while the second stage uses the long-context ICL reasoning subset to improve evidence grounding and adaptation to in-context information.

\textbf{Training Stages.} We adopt GRPO~\citep{shao2024deepseekmath} in both stages. For each prompt, the model generates a group of candidate responses, and the reward differences within the group provide self-contrastive supervision for policy optimization. To improve the efficiency of RL training, we apply dynamic sampling during rollout. Specifically, for each prompt group, we retain only groups with non-uniform rewards and filter out groups where all sampled responses receive the same reward. This strategy removes prompts that are either too easy or too hard for the current policy, since such groups provide little useful preference signal. As a result, the policy update is concentrated on samples with meaningful reward contrast. The RL training pipeline consists of two consecutive stages:
\begin{itemize}
    \item \textbf{Stage 1: Instruction-following RL.} 
    We first train the model on Nemotron instruction-following data~\citep{nemo-gym}. This stage aims to improve the model's ability to understand and satisfy fine-grained user constraints, such as formatting requirements, length limits, keyword inclusion or exclusion, language constraints, and other explicit instructions.  We use verifiable rule-based rewards. The reward function checks whether the generated response satisfies explicit constraints specified in the prompt, including formatting, length, keyword, language, and other rule-based requirements. This design provides a reliable supervision signal for instruction adherence while avoiding dependence on external judges.

    \item \textbf{Stage 2: Long-context Learning RL.}
    Starting from the instruction-following RL checkpoint, we continue training the model on long-context learning data. We use rule-based answer matching as the outcome reward. A response is rewarded when its final answer matches the ground-truth answer according to predefined matching rules. This stage encourages the model to ground its reasoning in task-specific information from the long input. By training on our constructed data, the model learns to retrieve sparse but decisive evidence, integrate information across distant document locations, adapt to newly introduced in-context rules, and reject salient but unsupported distractors. Consequently, the model relies less on memorized priors or local pattern matching, leading to stronger long-context learning behavior.
\end{itemize}


\subsubsection{Reinforcement Learning on Tool-calling}
For agentic tool use, SFT alone is insufficient because the main failure modes are not only local formatting errors. The agent must learn when to call tools, which tool to call, how to produce valid arguments, how to recover from tool errors, and when the task is actually complete. These decisions create delayed consequences: a wrong early tool call may only become visible many turns later, and premature stopping may look fluent while failing the task. RL is therefore required to optimize complete trajectories rather than isolated assistant turns.

\textbf{Reward Design and Advantage Enhancement.} To enable the model to obtain denser marginal signals and maximize marginal gains, we construct two types of reward functions. The first is an outcome reward indicating complete task success. The second is a process score measuring partial completion over LLM rubrics:
\begin{equation}
r_i^{\mathrm{out}} \in \{0,1\},
\qquad
r_i^{\mathrm{proc}}
=
\frac{1}{|\mathcal{R}|}
\sum_{j=1}^{|\mathcal{R}|}
\mathbf{1}[\mathcal{R}_j \text{ is satisfied}].
\end{equation}
The advantage uses an asymmetric design~\citep{tan2026papostabilizingrubricintegration}. Successful trajectories already satisfy all rubrics, so adding process reward to positive samples would double-count the same signal. The informative use of process reward is on failed trajectories, where different failures can be meaningfully ranked by how close they were to success. For group-based RL, the shaped advantage is: 
\begin{equation}
A_i
=
A_i^{\mathrm{out}}
+
\lambda_{\mathrm{neg}}
\mathbf{1}[r_i^{\mathrm{out}}=0]
A_i^{\mathrm{proc}},
\qquad
\lambda_{\mathrm{neg}}=0.5,
\end{equation}
where $A^{\mathrm{out}}$ is normalized over valid samples in the group, while $A^{\mathrm{proc}}$ is normalized only over valid negative samples. This preserves the relative ranking among failed trajectories and avoids using positive samples to distort the failure-side process distribution.

\textbf{Training Data.} The data strategy is designed around the general problem of sparse, noisy, long-horizon agent rewards. One component is a hard-task set, which provides challenging prompts where the SFT model has a low pass rate and many trajectories are near misses. This is useful for RL because it creates gradient-bearing contrast within a group. 
In RL post-training, data reuse is an effective strategy that repeatedly leverages the same batch of high-quality data, achieving an effect comparable to using a much larger batch of data within certain constraints. The hard-data component is intentionally reused rather than treated as a large one-pass corpus. For the training set has $|\mathcal{D}|$ tasks, rollout batch size $B$, samples per prompt $K$, and $R$ rollout rounds, the expected number of generated trajectories per task is approximately:
\begin{equation}
N_{\mathrm{reuse}}
\approx
\frac{R \cdot B \cdot K}{|\mathcal{D}|}.
\end{equation}

\textbf{Training Stages.} Our training pipeline consists of two stages: Tool-specific SFT and Tool-enhanced RL.
\begin{itemize}
    \item \textbf{Tool-specific SFT.} In the SFT stage, we use the data collected in Section \ref{sec:tool-calling} to enhance the tool-use capabilities of on Qwen3.5-35B-A3B~\citep{Qwen3.5}. The primary objective of this stage is to improve the model’s ability to generate tool-use instructions, produce correctly formatted tool calls, and perform basic tool invocation. Meanwhile, we employ a rubric model to record and monitor challenging tasks encountered at this stage. These tasks are then used for RL-based enhancement in the subsequent stage.
    \item \textbf{Tool-specific RL.} In the RL stage, we re-evaluate the trajectories of the aforementioned tasks using rubrics. This additional round of evaluation further filters the data and retains the higher-quality subset. A salient property of this subset is that it consists of near-success cases that nevertheless fail to obtain outcome rewards. Based on this process, we construct a hard-task set containing only 64 samples. By data reuse as described above and applying a PAPO-style advantage to enhance GRPO, we achieve efficient Tool RL improvement with a small amount of high-quality data over only a few training steps. 
\end{itemize}


\subsection{Multi-teacher On-Policy Distillation}
\label{sec:multi-opd}

After full-domain SFT and domain-level teacher training, we consolidate specialized teachers into a single deployable student through multi-teacher OPD. The student is optimized on its own rollouts under domain-specific teacher guidance, while the detailed OPD objective, including salient vocabulary alignment and domain-normalized aggregation, is introduced in Sec.~\ref{subsec:on-policy-distillation}. This section describes the training pipeline.

\textbf{Student initialization and teacher pool.}
The student is initialized from the full-domain SFT checkpoint in Sec.~\ref{sec:sft}. The teacher pool is built from the domain-level models in Sec.~\ref{sec:domain-level}, where each teacher is specialized through targeted SFT or RL. During OPD, teachers are not merged at the parameter level. Instead, each prompt is assigned a domain label, and the corresponding teacher provides the distillation signal for the student rollout.

\textbf{Training data and domain routing.}
\textbf{1) Data organization:} The OPD training set is reorganized from earlier task families for on-policy learning. Each example contains the user prompt, domain label, applicable interaction protocol, and execution metadata such as environment configuration, finalization rules, and verifier information. 
\textbf{2) Domain routing:} We deduplicate prompts and balance the number of unique prompts per domain. During training, each sample is routed to the teacher trained for its domain, preserving domain-specific preferences and avoiding incompatible teacher signals.

\textbf{On-policy rollout generation.}
\textbf{1) Rollout construction:} For each batch, the current student generates responses or agentic trajectories under the corresponding domain protocol. Tool outputs, user turns, and environment observations are kept as context but masked from the loss, so optimization applies only to student-generated tokens. 
\textbf{2) Rollout bounding:} To control the large variance in rollout structure, each rollout is bounded by a turn budget $T_{\max}$, response-length budget $L_{\max}^{\mathrm{resp}}$, and context-length budget $L_{\max}^{\mathrm{ctx}}$. Capped rollouts are marked \texttt{TRUNCATED} and retained as valid prefixes, while system-interrupted rollouts are marked \texttt{ABORTED} and retried.

\textbf{Teacher-guided policy optimization.}
After rollout collection, the routed teacher evaluates the student-generated prefixes and provides token-level guidance. Unlike offline imitation, the teacher does not generate a separate reference trajectory; it scores the student's own trajectory, making the signal on-policy. In our final configuration, the student is trained with the domain-routed SVA objective in Sec.~\ref{subsec:on-policy-distillation}, where losses are aggregated with domain-normalized weighting to balance heterogeneous teachers. Through OPD, the student retains broad SFT coverage while absorbing stronger domain-specific behaviors from the teacher pool into a unified long-horizon agent.

\section{Experimental Results}
\label{sec:exp}

\subsection{Evaluation Setting} 
\textbf{For the long-horizon search evaluation}, we cover four public benchmarks: GAIA~\citep{gaia}, BrowseComp~\citep{browsecomp}, XBench-DeepResearch~\citep{xbench}, and SEAL‑0~\citep{sealqa}. Each agent is equipped with three tools: a search tool that retrieves web pages and returns the top‑50 results per query; a visit tool that fetches webpage content and summarizes it with a dedicated summarization model to extract task‑relevant information; and a code tool that executes Python scripts in a remote sandbox to support complex computation and logical reasoning. We cap each task at 300 turns and report pass@1 as the primary metric. For answer verification, we strictly follow each benchmark's official judge model and prompt settings, rather than using a unified judge.

\textbf{For engineering tasks}, we evaluate SciCode~\citep{scicode} and MLE-Bench-Lite~\citep{mlebench} under their respective official protocols. SciCode targets research-level scientific coding tasks in which problems are decomposed into sequential subproblems, and solutions are judged correct only when they pass all associated hidden unit tests. Following the standard setup, we supply the scientist-annotated background for each subproblem and report pass@1 over the $288$ subproblems in the test set. MLE-Bench-Lite instead measures end-to-end machine-learning engineering on $22$ Kaggle competitions: given only a dataset and a task description, the agent has to autonomously explore the data, train models, and emit a submission, which is graded against the original competition leaderboard and mapped to a Kaggle medal (bronze, silver, or gold). We follow the official MLE-Bench grading and report the medal rate, \textit{i.e.}, the fraction of competitions in which the submission earns at least a bronze medal, averaged over three seeds. Every task runs in isolation on a dedicated H200 GPU with a 12-hour wall-clock budget.

\textbf{For scientific research evaluation}, we include four representative benchmarks: HLE with tools~\citep{phan2025humanity}, HiPhO~\citep{hipho}, FS-O (FrontierScience-Olympiad)~\citep{frontierscience}, and FS-R (FrontierScience-Research)~\citep{frontierscience}. HLE with tools evaluates expert-level reasoning with external tool use and is one of the most widely used public leaderboards for frontier reasoning. We report official scores for baseline models whenever available; for Qwen3.6-35B-A3B, which lacks an official result, we use the same tool-augmented pipeline as our model, including search, visit, code, and scholar tools. HiPhO is the first benchmark dedicated to physics Olympiad evaluation, assessing multimodal physics reasoning across $13$ recent competitions from 2024--2025, and we report the average score over all competitions. FS-O and FS-R evaluate olympiad-level and research-level scientific reasoning, respectively, across disciplines such as physics, chemistry, and biology. We follow the official protocols and report average accuracy. For HiPhO, FS-O, and FS-R, we report tool-free results for comparison models to avoid confounding effects from applying our tool-augmented evaluation protocol to models not trained for tool use, which can in some cases degrade performance. This setup also provides a stringent comparison, as it tests whether our tool-equipped agentic model can outperform leading large-scale frontier models under their standard evaluation configurations.

\textbf{For long-context and instruction-following evaluation}, we use LongBench V2~\citep{LongBench}, IFBench~\citep{IFBench}, and IFEval~\citep{zhou2023ifeval}under the evaluation protocols implemented in their official or benchmark-provided scripts. LongBench V2 evaluates long-context understanding over 503 multiple-choice questions spanning single-document QA, multi-document QA, long in-context learning, long-dialogue history understanding, code repository understanding, and long structured data understanding. We use the chain-of-thought prompting setting, truncate inputs to 128K tokens when necessary, and extract the final multiple-choice answer, reporting accuracy over the full set as well as breakdowns by difficulty and context length. IFBench and IFEval measure fine-grained instruction-following ability. The model generates a response for each prompt, and rule-based validators check whether all specified constraints are satisfied. IFBench contains 294 test prompts and IFEval contains 541 prompts. For both benchmarks, we follow the benchmark evaluation scripts and report strict instruction-following accuracy at both the prompt level and instruction level.

\textbf{For general agentic tasks}, both $\tau^2$-Bench~\citep{barres2025tau2} and VitaBench~\citep{he2025vitabench} use pass@1 averaged over all domains. Specifically, $\tau^2$-Bench covers retail, telecom, and airline, while VitaBench covers cross-domain, delivery, in-store, and OTA. The official settings of both benchmarks use GPT-4.1 as the user simulator; to reduce potential reproducibility risks from future model retirement, we use the open-source DeepSeek-V3.2~\citep{liu2025deepseek} as the user simulator for both benchmarks. For VitaBench, we also use DeepSeek-V3.2 as the judge, replacing the Claude-3.7-Sonnet judge used in the original report (Anthropic states that Claude-3.7-Sonnet has been retired and is no longer available~\footnote{\url{https://platform.claude.com/docs/en/about-claude/model-deprecations}}). 

For MolBench~\citep{zhang2026molclaw}, we report the score of Binding Affinity Comparison (MolBench-bind.), following the official evaluation setting and averaging over three repeated runs. For MatTools~\citep{MatTools}, we use an autonomous code-exploration setting: the model is allowed to inspect, interact with, and explore the nearly 100K-line tool codebase over multiple turns before completing the tasks. We report the completion rate over 138 subtasks, averaged over three runs. For inference settings, GPT-5.5 uses xhigh reasoning effort, while all other models use temperature 0.7 with other parameters kept at their default values.

\subsection{Results and Observations} 

\begin{table}[]
    \centering
    \footnotesize
    \setlength{\tabcolsep}{1.9pt}
    \caption{Performance comparison of Qwen3.5-35B-A3B, \ProjectName-SFT, and \ProjectName.}
    \label{tab:sft_opd}
    \begin{tabular}{@{}lccc@{}}
    \toprule
    & \multicolumn{1}{l}{\textbf{Qwen3.5-35B-A3B}} & \multicolumn{1}{l}{\textbf{\ProjectName-SFT}} & \multicolumn{1}{l}{\textbf{\ProjectName}} \\
    \midrule
    \textbf{Long-horizon Search} & & & \\
    \midrule
    BrowseComp~\citep{browsecomp} & 61.0 & 74.6 & \textbf{75.5} \\
    XBench-DS-2510~\citep{xbench} & 77.0 & \textbf{88.0} & 86.0 \\
    Seal-0~\citep{sealqa} & 41.4 & 52.3 & \textbf{56.4} \\
    GAIA~\citep{gaia} & 59.8 & 95.2 & \textbf{96.0} \\
    \midrule
    \textbf{Engineering Tasks} & & & \\
    \midrule
    SciCode~\citep{scicode} & 37.1 & 42.3 & \textbf{44.3} \\
    MLE-Bench-Lite~\citep{mlebench} & 24.2 & 39.4 & \textbf{43.9} \\
    \midrule
    \textbf{Scientific Research} & & & \\
    \midrule
    HLE w/ tools~\citep{phan2025humanity} & 47.4 & 41.6 & \textbf{47.6} \\
    HiPhO~\citep{hipho} & 37.0 & 42.9 & \textbf{46.4} \\
    FS-O~\citep{frontierscience} & 64.5 & 75.0 & \textbf{79.0} \\
    FS-R~\citep{frontierscience} & 2.5 & 31.7 & \textbf{40.0} \\
    \midrule
    \textbf{Instruction Following} & & & \\
    \midrule
    IFbench~\citep{IFBench} & 70.2 & 68.7 & \textbf{80.6} \\
    Longbench V2~\citep{LongBench} & 59.0 & 58.3 & \textbf{60.2} \\
    \midrule
    \textbf{General Agentic Tasks} & & & \\
    \midrule
    $\tau^2$-Bench~\citep{barres2025tau2} & \textbf{81.2} (32.5)$^\dag$ & 76.7 & 79.8 \\
    VitaBench~\citep{he2025vitabench} & 26.0 & 37.3 & \textbf{38.8} \\
    \midrule
    \textbf{Scientific Agentic Tasks} & & & \\
    \midrule
    MatTools~\citep{MatTools} & 21.0 & 37.0 & \textbf{47.1} \\
    MolBench-Bind.~\citep{zhang2026molclaw} & 46.0 & 46.0 & \textbf{56.8} \\
    \bottomrule
    \end{tabular}
    
    \vspace{1mm}
    \footnotesize{$^\dag$ For $\tau^2$-Bench, we report both the official Qwen3.5-35B-A3B result (81.2) and our reproduced result (33.0); see Section~\ref{sec:sft_results} for discussion.}
\end{table}

\subsubsection{Full-domain SFT Results} 
\label{sec:sft_results}

The results of the SFT-stage model are shown in Table~\ref{tab:sft_opd}. From the results, it can be observed that compared with Qwen3.5-35B-A3B, \ProjectName-SFT shows clear improvements on long-horizon search, engineering tasks, scientific research, and agentic tasks. However, we also observe that \ProjectName-SFT has clear performance drops on general agentic tasks, instruction following, and HLE. \textit{We believe this is mainly caused by the difference between the long-thinking reasoning pattern and the multi-turn agentic pattern. Full-domain SFT cannot easily solve the domain conflicts caused by different reasoning patterns.} Therefore, we choose the multi-domain on-policy distillation (OPD). Before presenting the experimental results of OPD, we show the results of the teacher model training for each domain.

\subsubsection{Results of Domain Teacher Training} 
\textbf{Experiments on Search Tasks.} As shown in Table~\ref{tab:search_rl_results}, the search-enhanced teacher consistently outperforms the Qwen3.5-35B-A3B model across all four benchmarks. The most notable improvement is observed on GAIA, where the score increases from 59.8 to 85.4 (+25.6). On HLE, the search-enhanced teacher yields a moderate improvement of 2.9 points (47.4 $\to$ 50.3).

\begin{table}[]
    \centering
    \footnotesize
    \setlength{\tabcolsep}{8pt}
    \renewcommand{\arraystretch}{1.4}
    \caption{Performance comparison between Qwen3.5-35B-A3B and the search-enhanced Teacher.}
    \label{tab:search_rl_results}

    \begin{tabular}{@{}lcccc@{}}
    \toprule
    \textbf{Model} & \textbf{GAIA} & \textbf{Seal-0} & \textbf{HLE w/ tools} & \textbf{XBench-DS-2510} \\
    \midrule
    Qwen3.5-35B-A3B & 59.8 & 41.4 & 47.4 & 77.0 \\
    Search-enhanced Teacher (SFT+RL) & \textbf{95.1} & \textbf{54.1} & \textbf{50.3} & \textbf{86.0} \\
    \bottomrule
    \end{tabular}
\end{table}

\textbf{Experiments on Scientific Domain.} It can be observed from Table~\ref{tab:sci-enhanced-teacher} that the science-enhanced teacher displays a comprehensive superiority compared with the baseline Qwen3.5-35B-A3B, especially on FS-R. It demonstrates that the proposed two-stage SFT can significantly promote both the intrinsic reasoning ability and extrinsic tool-use interaction capability in scientific scenarios. 

\begin{table*}[]
\footnotesize
\centering
\setlength{\tabcolsep}{8pt}
\renewcommand{\arraystretch}{1.4}
\caption{Performance comparison between Qwen3.5-35B-A3B and the science-enhanced teacher. Tool usage is allowed for all benchmarks.}
\label{tab:sci-enhanced-teacher}
\begin{tabular}{lcccc}
\midrule
Model & HLE w/ tools  & HiPhO & FS-O & FS-R \\ 
\midrule
Qwen3.5-35B-A3B & 47.4 & 37.0  & 64.5 & 2.5  \\
Science-enhanced Teacher (SFT) & \textbf{47.8} & \textbf{46.9} & \textbf{82.0} & \textbf{54.3} \\ 
\midrule
\end{tabular}
\end{table*}
\begin{table*}[t]
\footnotesize
\centering
\setlength{\tabcolsep}{4pt}
\renewcommand{\arraystretch}{1.3}
\caption{Evaluation results of Qwen3.5-35B-A3B and RL-enhanced teacher on LongBench V2, IFBench, and IFEval, where IFBench and IFEval report strict scores.}
\begin{tabular}{lccc}
\toprule
\textbf{Model}
& \textbf{LongBench V2}
& \textbf{IFBench}
& \textbf{IFEval} \\
\midrule

Qwen3.5-35B-A3B
& 59.0
& 70.2
& 91.9 \\

Long-instruction-enhanced RL
& \textbf{62.4}
& \textbf{82.0}
& \textbf{93.4} \\
\bottomrule
\end{tabular}
\label{tab:long_sft_rl_results}
\end{table*}

\paragraph{Experiments on Instruction Following and Long-Context Learning}
As shown in Table~\ref{tab:long_sft_rl_results}, the RL-enhanced teacher consistently improves over Qwen3.5-35B-A3B on both long-context and instruction-following evaluations. On LongBench v2, the overall score increases from 59.0 to 62.4. This suggests that our long-context RL stage mainly strengthens the model's ability to retrieve and understand relevant evidence from longer and more difficult contexts, thereby improving complex long-context learning. On IFBench, the strict score increases from 70.2 to 82.0, indicating stronger generalization to challenging verifiable instruction constraints. The model also improves on IFEval. Overall, these results show that our RL enhancement effectively improves long-context learning and precise instruction-following capability.

\textbf{Experiments on Tool-calling.} We further evaluate the effect of tool-enhanced post-training on tool-calling benchmarks. As shown in Table~\ref{tab:tool_rl_results}, the tool-enhanced model, obtained by applying SFT and RL on top of Qwen3.5-35B-A3B, brings substantial improvements on $\tau^2$-Bench and VitaBench. On $\tau^2$-Bench, the average score increases from 32.53 to 82.50, with consistent gains on Airline, Retail, and Telecom. In particular, Airline improves from 16.00 to 72.00, and Retail improves from 30.70 to 82.50, suggesting that tool-enhanced post-training strengthens the model's ability to follow domain-specific tool-use constraints and complete multi-turn operational tasks. On VitaBench, the average score improves from 26.00 to 44.16. These results indicate that tool-calling ability benefits strongly from explicit tool-use supervision and reinforcement learning, especially when the tasks require structured interaction with external environments rather than pure language understanding. 

\begin{table*}[h]
\footnotesize
\centering
\setlength{\tabcolsep}{4pt}
\renewcommand{\arraystretch}{1.3}
\caption{Evaluation results of Qwen3.5-35B-A3B and the tool-enhanced RL teacher on $\tau^2$-bench and Vita-bench.}
\begin{tabular}{lccccccccc}
\toprule
& \multicolumn{4}{c}{$\tau^2$-bench\textsuperscript{\dag}} 
& \multicolumn{5}{c}{VitaBench} \\
\cmidrule(lr){2-5} \cmidrule(lr){6-10}
\textbf{Model}
& \textbf{Airline}
& \textbf{Retail}
& \textbf{Telecom}
& \textbf{Avg}
& \textbf{Cross Domain}
& \textbf{Delivery}
& \textbf{In-store}
& \textbf{Ota}
& \textbf{Avg} \\
\midrule
Qwen3.5-35B-A3B
& 16.00
& 30.70
& 50.90
& 32.53
& 11.51
& 39.25
& 31.50
& 21.75
& 26.00 \\

Tool-enhanced (SFT+RL)
& \textbf{72.00}
& \textbf{82.50}
& \textbf{93.00}
& \textbf{82.50}
& \textbf{30.00}
& \textbf{56.00}
& \textbf{51.75}
& \textbf{38.89}
& \textbf{44.16} \\
\bottomrule
\multicolumn{10}{l}{\footnotesize \textsuperscript{\dag} For $\tau^2$-Bench, we report our reproduced result (32.53); see Section~\ref{sec:sft_results} for discussion.} \\
\end{tabular}
\label{tab:tool_rl_results}
\end{table*}

\subsubsection{Results of On-policy Distillation} 

\begin{table}[]
    \centering
    \footnotesize
    \setlength{\tabcolsep}{1.9pt}
    \caption{Comparison between \ProjectName~ and 35B/1T-level models, where FS-O, FS-R, and MolBench-Bind represent FrontierScience-Olympiad~\citep{frontierscience}, FrontierScience-Research~\citep{frontierscience}, and MolBench-binding affinity comparison~\citep{zhang2026molclaw}. For 35B model, we compare with recently released open-source 35B models. For 1T-level models, we compare with Kimi-K2.6~\citep{kimi2.6}, DeepSeek-V4-Pro~\citep{deepseekai2026deepseekv4} with Max reasoning effort, and GPT-5.5~\citep{GPT5.5}. To ensure a fair comparison, we report the results from their original technical reports. If a model does not report the corresponding benchmark results, we evaluate it using the same evaluation protocol as our model. \underline{underline} means the best result for 35B parameters, and \textbf{Bold} means the best overall result. }
    \label{tab:main_results}

    \begin{tabular}{@{}lcccccccc@{}}
    \toprule
    & \multicolumn{4}{c}{\textbf{35B parameters}} & \multicolumn{3}{c}{\textbf{\textgreater{}1T parameters}} \\
    \cmidrule(lr){2-4} \cmidrule(lr){5-7}
    & \multicolumn{1}{l}{\textbf{\ProjectName}} & \multicolumn{1}{l}{\textbf{Qwen3.6-35B-A3B}} & \multicolumn{1}{l}{\textbf{Nex-N2-mini}} & \multicolumn{1}{l}{\textbf{Kimi-K2.6}} & \multicolumn{1}{l}{\textbf{DSV4-Pro (Max)}} & \multicolumn{1}{l}{\textbf{GPT-5.5}} \\
    \midrule
    \textbf{Long-horizon Search} & \multicolumn{1}{l}{} & \multicolumn{1}{l}{} & \multicolumn{1}{l}{} & \multicolumn{1}{l}{} & \multicolumn{1}{l}{} & \multicolumn{1}{l}{} & \multicolumn{1}{l}{} & \multicolumn{1}{l}{} \\
    \midrule
    BrowseComp~\citep{browsecomp} & \underline{75.5} & 67.9 & 74.1 & 83.2 & 83.4 & \textbf{84.4} \\
    XBench-DS-2510~\citep{xbench} & \underline{86.0} & 71.0 & 82.0 & \textbf{90.0} & \textbf{90.0} & 84.0 \\
    Seal-0~\citep{sealqa} & \textbf{56.4} & 38.7 & 49.6 & 50.5 & 55.0 & 42.3 \\
    GAIA~\citep{gaia} & \underline{96.0} & 78.6 & 82.5 & 80.6 & \textbf{98.1} & 87.4 \\
    \midrule
    \textbf{Engineering Tasks} & & & & & & \\
    \midrule
    SciCode~\citep{scicode} & \underline{44.3} & 35.8 & 29.9 & 53.5 & 50.0 & \textbf{56.1} \\
    MLE-Bench-Lite~\citep{mlebench} & \underline{43.9} & 34.9 & 34.9 & 62.1 & 63.6 & \textbf{72.7} \\
    \midrule
    \textbf{Scientific Research} & & & & & & \\
    \midrule
    HLE w/ tools~\citep{phan2025humanity} & \underline{47.6} & 36.2 & 32.0 & \textbf{54.0} & 48.2 & 52.2 \\
    HiPhO~\citep{hipho} & \textbf{46.4} & 37.7 & 38.5 & 41.1 & 38.7 & 43.3 \\
    FS-O~\citep{frontierscience} & \textbf{79.0} & 60.3 & 52.0 & 73.0 & 76.0 & 78.0 \\
    FS-R~\citep{frontierscience} & \textbf{40.0} & 2.9 & 5.0 & 17.9 & 13.3 & 26.7 \\
    \midrule
    \textbf{Instruction Following} & & & & & & \\
    \midrule
    IFbench~\citep{IFBench} & \textbf{80.6} & 64.4 & 54.1 & 71.8 & 73.5 & 75.9 \\
    Longbench V2~\citep{LongBench} & \underline{60.2} & 57.7 & 59.6 & 62.0 & \textbf{64.3} & -\\
    \midrule
    \textbf{General Agentic Tasks} & & & & & & \\
    \midrule
    $\tau^2$-Bench~\citep{barres2025tau2} & \underline{79.8} & 79.0 & 74.5 & 81.9 & \textbf{82.2} & 81.6 \\
    VitaBench~\citep{he2025vitabench} & \underline{38.8} & 35.6 & 23.0 & 35.6 & \textbf{49.0} & 45.0 \\
    \midrule
    \textbf{Scientific Agentic Tasks} & & & & & & \\
    \midrule
    MatTools~\citep{MatTools} & \underline{47.1} & 15.9 & 34.1 & 63.8 & 47.1 & \textbf{68.8} \\
    MolBench-Bind.~\citep{zhang2026molclaw} & \underline{56.8} & 48.7 & 51.4 & 21.6 & 37.8 & \textbf{62.2} \\
    \bottomrule
    \end{tabular}
\end{table}

The results of Multi-Domain-Routed On-Policy Distillation based on the \ProjectName-SFT model are shown in Table~\ref{tab:main_results}. It can be observed from Table~\ref{tab:main_results} that \ProjectName~is a strong 35B-level model and can compete with much larger 1T-level models on many difficult tasks. \ProjectName~outperforms same-scale 35B baselines and even surpassing several 1T-level models in multi-step search, scientific research, and long-instruction following. We find that the abilities of multi-step search, scientific research, and long-instruction following can support each other. For example, when solving complex scientific problems, improving the model’s ability to use external tools, such as search tools or code tools, helps the model choose the right tool in open-ended tasks and get external knowledge more efficiently. This further improves its performance on scientific research tasks.

Besides, \ProjectName~achieves 44.3 on SciCode and 43.9 on MLE-Bench-Lite, leading all same-scale 35B baselines on both benchmarks. However, it is still clearly weaker than 1T-level models. For example, GPT-5.5 reaches 72.7 on MLE-Bench-Lite. We believe this is mainly due to that MLE optimization is not a static problem-solving task. Instead, it requires the model to complete a full engineering process. This places higher demands on keeping a stable goal, remembering past decisions, and avoiding repeated trials across many experiments.

For $\tau^2$-Bench, Qwen3.5-35B-A3B baseline we reproduced only gets a score below 40. Some other community works also report a similar issue~\footnote{\url{https://github.com/thinking-machines-lab/tinker-cookbook/blob/main/tinker_cookbook/eval}}. We think this discrepancy may stem from differences across $\tau^2$-Bench codebase versions and evaluation environments, a topic that has also been discussed within the community~\footnote{\url{https://github.com/QwenLM/Qwen3/discussions/1809}}.
On Vita, MatTools, and MolBench-MS Binding Affinity tasks, we observe consistent improvements from \ProjectName.

Besides, Table~\ref{tab:sft_opd} shows the performance comparison between \ProjectName-SFT and \ProjectName~(trained with OPD). We can observe that the thinking pattern of long-instruction following is very different from that of long-horizon search. The former uses a single-turn and long-thinking pattern, while the latter uses a multi-turn tool-use and short-thinking pattern. The performance drop in the full-domain SFT stage, caused by different thinking patterns, can be substantially reduced by the Multi-teacher Multi-domain OPD stage. 

By comparing Table~\ref{tab:sci-enhanced-teacher}, Table~\ref{tab:long_sft_rl_results} and Table~\ref{tab:main_results}, we observe that the OPD-trained model does not always outperform the corresponding domain teacher. This is expected, since each teacher is specialized for one domain, while \ProjectName~is required to maintain a unified policy across heterogeneous tasks. In our empirical setting, OPD mainly serves to transfer teacher strengths into a single model and improve the balance over full-domain SFT, rather than consistently exceeding every teacher on its own specialty.

Based on the above experimental observations, we hope that \ProjectName~can provide the community with a clear technical path to unify more diverse agentic scenarios and tasks. Besides, we pay special attention to the model’s performance on long-horizon agentic tasks. Therefore, we analyze the strengths and limitations of \ProjectName~through several long-horizon cases in the following section.

\subsection{Long-Horizon Task Applications}

\subsubsection{A 12-Hour Long-Horizon Optimization Run}
\label{mle:example}
\vspace{-0.5em}

To evaluate the multi-step reasoning capability of \ProjectName~on machine learning engineering tasks, we select the right whale call detection task from the MLE dataset as a representative subtask~\cite{mlebench} and require the model to perform end-to-end optimization over a long-horizon run. In this case, \ProjectName~starts from a naive CNN baseline and autonomously improves the pipeline through a sequence of selected interventions, including temporal data analysis, audio augmentation, temporally localized training, architectural refinement with Mel-spectrogram CNN ensembles, and large-scale augmentation. Across a 12-hour optimization trajectory, the model progressively raises the best validation AUC from 0.58 to 0.9935, ultimately reaching a gold-medal-level result.

\begin{figure}[t]
\centering
\includegraphics[width=\textwidth]{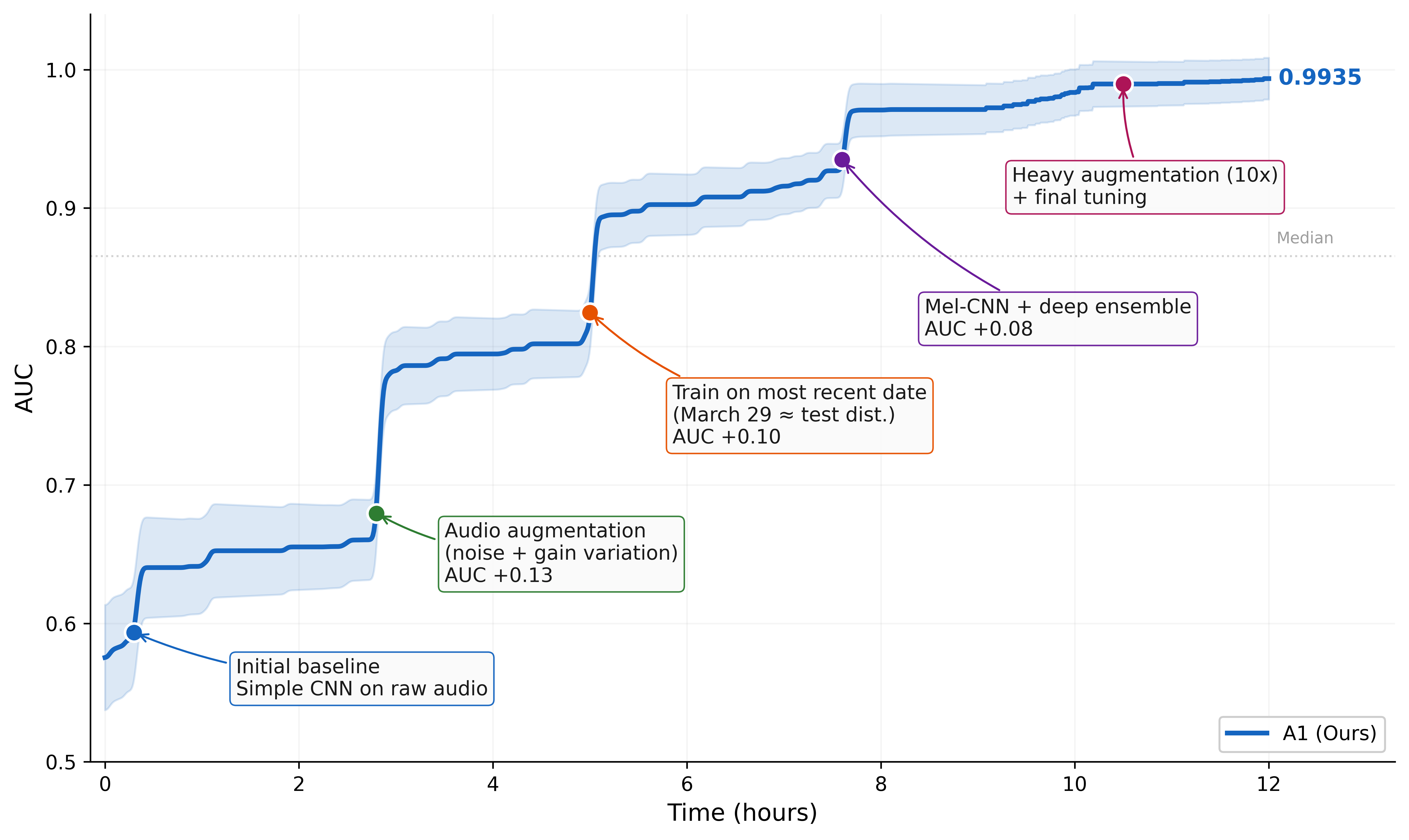}
\caption{Optimization trajectory of \ProjectName~on the ICML 2013 Whale Challenge~\cite{icml2013bioacoustics_challenge} over a 12-hour run. The curve shows the best validation AUC achieved over wall-clock time, with annotated breakthrough moments corresponding to distinct algorithmic improvements. The shaded band indicates run-to-run variance across independent seeds.}
\label{fig:mle_example}
\end{figure}

\begin{figure}[t]
\centering
\includegraphics[width=\textwidth]{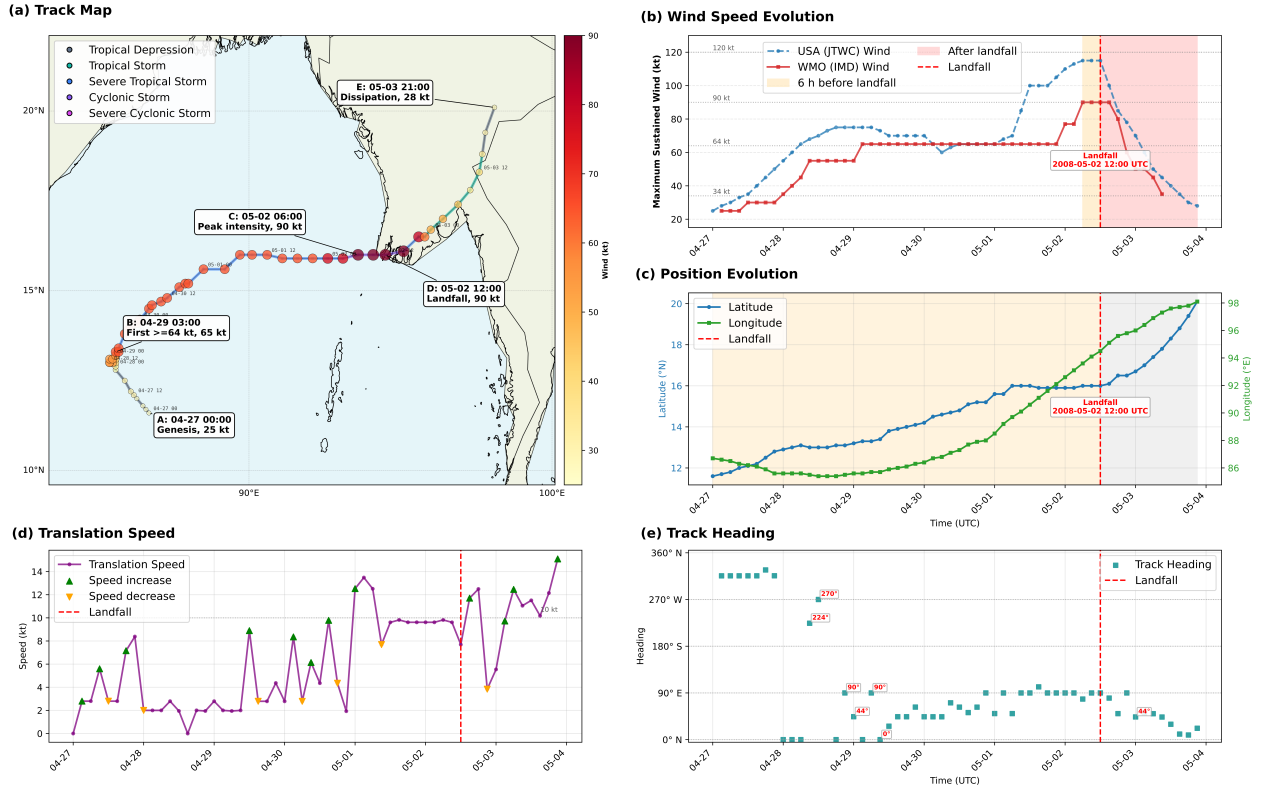}
\caption{Integrated track, intensity, and motion characteristics of Tropical Cyclone Nargis (2008) produced by \ProjectName. (a) Best-track map of Nargis, with track-segment colors indicating intensity class and marker colors indicating maximum sustained wind speed; key stages are labeled A--E. (b) Temporal evolution of maximum sustained wind speed from USA/JTWC and WMO/IMD estimates, with orange shading indicating the 6~h pre-landfall period, red shading indicating the post-landfall period, and the red dashed line denoting landfall at 12:00 UTC on 2 May 2008. (c) Temporal evolution of latitude and longitude, shown on the left and right y-axes, respectively. (d) Translation speed as a function of time, with triangles marking significant acceleration and deceleration events. (e) Track heading as a function of time, with red annotations indicating rapid turning events. All times are in UTC.}
\label{fig:earth_science_nargis}
\end{figure}

This example shows that \ProjectName~can perform long-horizon model optimization beyond isolated hyperparameter tuning. Figure~\ref{fig:mle_example} reflects a consistent improvement direction across multiple iterations, identifies a meaningful temporal domain shift between training and test recordings, and applies targeted algorithmic changes that substantially improve generalization performance. Taken together, these results indicate that \ProjectName~can integrate dataset diagnosis, representation design, augmentation strategy, and iterative evaluation to produce an interpretable multi-step optimization solution for a challenging real-world machine learning task.


\subsubsection{Closing the Loop in Earth Science Analysis with \ProjectName}
\label{earth_science:example}

To evaluate the end-to-end analytical capability of \ProjectName~on Earth science tasks, we select Severe Cyclonic Storm Nargis (2008) over the North Indian Ocean as a representative case and require the model to reconstruct the storm track, generate diagnostic visualizations, and interpret its track and intensity evolution from real best-track data. In this Earth science attempt, \ProjectName~automatically identifies IBTrACS as the data source~\cite{knapp2010ibtracs,gahtan2024ibtracs} and completes \textbf{data extraction}, \textbf{cleaning}, \textbf{derived-metric computation}, \textbf{visualization}, and \textbf{result synthesis}, forming a multi-stage closed loop of planning, coding, execution, result checking, scientific analysis, and report generation.

The results show that \ProjectName~reconstructs the major evolution of Nargis with reasonable fidelity, including its formation over the central Bay of Bengal, northwestward motion, later recurvature toward the east-northeast, and eventual landfall over southern Myanmar~\cite{jtwc2008atcr}. It further derives diagnostic quantities such as track length, translation speed, heading variation, and intensity evolution, while preserving both WMO/IMD and JTWC/USA intensity estimates to avoid conflating different operational conventions~\cite{ibtracs2025columns}. Figure~\ref{fig:earth_science_nargis} summarizes the key diagnostics of storm track, intensity, position, translation speed, and heading change, showing that this case provides an informative example of \ProjectName's capability for Earth science data organization, diagnostic computation, and result integration.

\section{Limitation and Future Work}
In this work, we have introduced a 35B MoE model \ProjectName. Our goal is to explore a promising technical path for building an agentic model by scaling the agent horizon. As an early effort toward scaling the agent horizon, the agentic abilities learned by our model mainly come from three sources: the initialization baseline ability of Qwen3.5-35B-A3B, the unified fundamental knowledge-action infrastructure built for different long-horizon scenarios, and a domain-routed on-policy distillation method that can reduce conflicts between reasoning patterns from different domains.

In our effort to scale the agent horizon from the baseline model, we also found that several basic atomic abilities are important for keeping the agent goal-consistent and efficient during long-horizon task-solving. These abilities include planning before reasoning, reflection before acting, summarizing key information in long contexts, and identifying important past information. In future work, we will focus on improving these basic atomic abilities for long-interaction agents, and use them as a starting point to further improve the ability of \ProjectName~to solve long-process tasks.

\begingroup
\sloppy
\printbibliography[heading=bibintoc]
\endgroup

\clearpage
\appendix
\section{Appendix}

\subsection{Contributions and Acknowledgments}

\definecolor{damaiblue}{RGB}{10, 102, 155}
\definecolor{damaiorange}{RGB}{180,50,50}
\definecolor{damaired}{RGB}{10, 50, 50}

\noindent
\textbf{\color{damaired} Knowledge-Action Infrastructure:} Zongsheng Cao\footnotemark[2]\footnotetext[2]{key contribution to this project}, Bihao Zhan, Zhijie Zhong

\noindent
\textbf{\color{damaired} Full-domain SFT:} Yue Fan\footnotemark[2], Tianshuo Peng

\noindent
\textbf{\color{damaired} Multi-teacher OPD:} Shiyang Feng\footnotemark[2], Yi Xie, Songtao Huang

\noindent
\textbf{\color{damaired} Long-horizon Search:} Tianshuo Peng\footnotemark[2], Zhijie Zhong, Jinxin Shi, Runmin Ma, Jiakang Yuan, Yusong Hu, Yue Fan

\noindent
\textbf{\color{damaired} Engineering Tasks:} Xiangchao Yan\footnotemark[2], Shangheng Du, Shuaiyu Zhang, Junpeng Zhao, Jinxin Shi, Yiming Wu, Boyuan Sun

\noindent
\textbf{\color{damaired} Scientific Research:} Fangchen Yu\footnotemark[2], Shengji Tang\footnotemark[2], Zhuo Liu, Jingqi Ye, Yichen Jiang, Haonan He, Weihao Lin

\noindent
\textbf{\color{damaired} Instruction Following and Context Learning:} Xiaohan He\footnotemark[2], Songtao Huang, Zhijie Zhong, Shiyang Feng

\noindent
\textbf{\color{damaired} General and Scientific Tool-calling:} Yiqun Zhang\footnotemark[2], Chen Zhang\footnotemark[2], Hao Li, Yang Chen, Chunjiang Mu, Zhiyao Cui, Qianyi Wang, Zelin Tan 

\noindent
\textbf{\color{damaired} Evaluation and Deployment:} Yuhao Zhou\footnotemark[2], Luohe Shi, Runmin Ma, Haoyang Peng, Wenjie Lou, Zijie Guo \\

\noindent
\textbf{\color{damaired} Scientific Directors and Advisors} \\  [1.5mm]
Wenlong Zhang, Fenghua Ling, Xin Li, Yan Teng, Dongrui Liu, Shufei Zhang, Liang He, Xiaosong Wang, Peng Ye, Shuyue Hu, Dahua Lin, Bowen Zhou

\vspace{12pt}
\noindent
\textbf{\color{damaired} Project Co-lead} \\ [1.5mm]
Bo Zhang, zhangbo@pjlab.org.cn \\
Lei Bai, bailei@pjlab.org.cn \\


\end{document}